\documentclass[conference]{IEEEtran}
\IEEEoverridecommandlockouts
\usepackage{cite}
\usepackage{amsmath,amssymb,amsfonts}
\usepackage{algorithmic}
\usepackage{booktabs}
\usepackage{booktabs}
\usepackage{graphicx}
\usepackage{algorithm}
\usepackage{textcomp}
\usepackage{multirow}
\usepackage{float}
\usepackage{subcaption}

\usepackage[most]{tcolorbox}
\usepackage{xcolor}
\usepackage{url}
\def\BibTeX{{\rm B\kern-.05em{\sc i\kern-.025em b}\kern-.08em
    T\kern-.1667em\lower.7ex\hbox{E}\kern-.125emX}}

\makeatletter
\newcommand{\linebreakand}{
  \end{@IEEEauthorhalign}
  \hfill\mbox{}\par
  \mbox{}\hfill\begin{@IEEEauthorhalign}
}
\makeatother
\begin{document}

\title{SHAP Distance: An Explainability-Aware Metric for Evaluating the Semantic Fidelity of Synthetic Tabular Data\\
}

\author{\IEEEauthorblockN{Ke Yu}
\IEEEauthorblockA{
\textit{The University of Tokyo}\\
Tokyo, Japan\\
yuke991001@g.ecc.u-tokyo.ac.jp}
\and
\IEEEauthorblockN{Shigeru Ishikura}
\IEEEauthorblockA{
\textit{Infomart Corporation}\\
Tokyo, Japan \\
s.ishikura@infomart.co.jp}
\and
\IEEEauthorblockN{Yukari Usukura}
\IEEEauthorblockA{
\textit{Infomart Corporation}\\
Tokyo, Japan \\
y.usukura@infomart.co.jp}
\linebreakand
\IEEEauthorblockN{Yuki Shigoku}
\IEEEauthorblockA{
\textit{Infomart Corporation}\\
Tokyo, Japan \\
y.shigoku@infomart.co.jp}
\and
\IEEEauthorblockN{Teruaki Hayashi}
\IEEEauthorblockA{
\textit{The University of Tokyo}\\
Tokyo, Japan\\
hayashi@sys.t.u-tokyo.ac.jp}
}

\maketitle

\begin{abstract}
Synthetic tabular data, which are widely used in domains such as healthcare, enterprise operations, and customer analytics, are increasingly evaluated to ensure that they preserve both privacy and utility. 
While existing evaluation practices typically focus on distributional similarity (e.g., the Kullback–Leibler divergence) or predictive performance (e.g., Train-on-Synthetic-Test-on-Real (TSTR) accuracy), these approaches fail to assess semantic fidelity, that is, whether models trained on synthetic data follow reasoning patterns consistent with those trained on real data. 
To address this gap, we introduce the SHapley Additive exPlanations (SHAP) Distance, a novel explainability-aware metric that is defined as the cosine distance between the global SHAP attribution vectors derived from classifiers trained on real versus synthetic datasets. 
By analyzing datasets that span clinical health records with physiological features, enterprise invoice transactions with heterogeneous scales, and telecom churn logs with mixed categorical–numerical attributes, we demonstrate that the SHAP Distance reliably identifies semantic discrepancies that are overlooked by standard statistical and predictive measures. 
In particular, our results show that the SHAP Distance captures feature importance shifts and underrepresented tail effects that the Kullback–Leibler divergence and Train-on-Synthetic-Test-on-Real accuracy fail to detect. 
This study positions the SHAP Distance as a practical and discriminative tool for auditing the semantic fidelity of synthetic tabular data, and offers practical guidelines for integrating attribution-based evaluation into future benchmarking pipelines.
\end{abstract}


\section{Introduction}
\begin{figure}[htbp]
    \centering
    \includegraphics[width=0.95\linewidth]{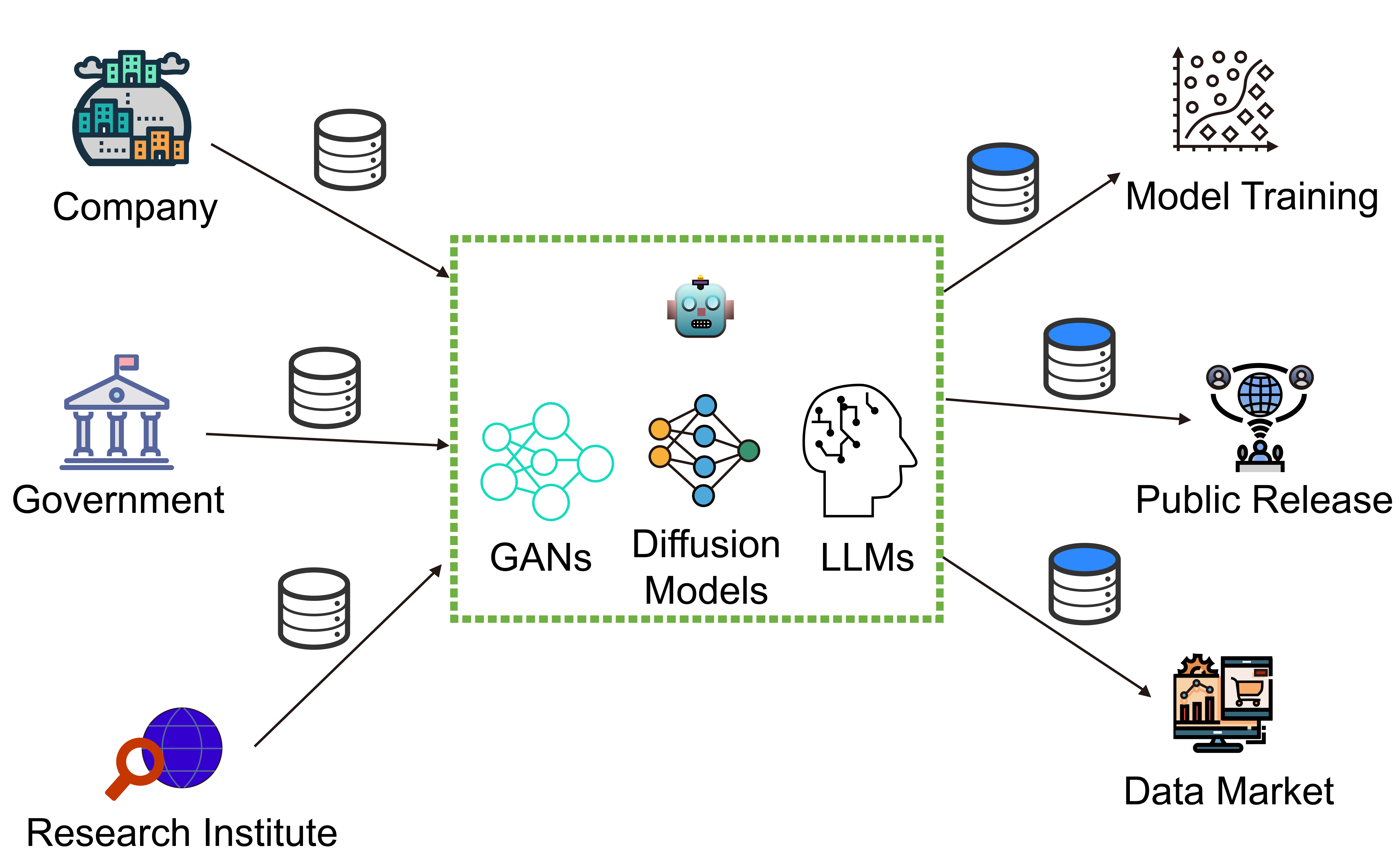}
    \caption{Overview of the synthetic data generation pipeline and downstream applications.}
    \label{fig:synthetic_pipeline_overview}
\end{figure}

Synthetic tabular data are increasingly recognized as a solution for enabling data sharing, model training, and system testing when access to real data is restricted by privacy or scarcity constraints. As shown in Figure~\ref{fig:synthetic_pipeline_overview}, synthetic data are generated from various sources using generative models, such as Generative Adversarial Networks (GANs), diffusion models, and Large Language Models (LLMs), and are deployed in tasks such as model training, public release, and data marketplaces. The breakthrough of GANs \cite{goodfellow2014gan} opened the door to data synthesis, and subsequent studies tailored GANs to tabular domains, such as conditional generators for mixed-type tables \cite{xu2019ctgan} and discrete electronic health records \cite{choi2017ehr}. More recently, diffusion-based models such as TabDDPM \cite{kotelnikov2023tabddpm} have shown improved mode coverage and sample diversity, underscoring the rapid progress of generative modeling.

In addition to new generators, surveys and frameworks highlight the importance of evaluation. Santangelo et al. emphasize that synthetic data must be assessed not only for privacy but also for utility \cite{santangelo2025synthro}, whereas Goncalves et al. present systematic evaluation protocols in the healthcare domain \cite{goncalves2020healthcare}. Toolkits such as SDMetrics \cite{sdv} provide standardized quality measures, yet most focus on statistical resemblance (e.g., the Kullback–Leibler divergence and pairwise correlation) or predictive performance in Train-on-Synthetic-Test-on-Real (TSTR) settings.

However, achieving statistical similarity does not guarantee semantic fidelity. A classifier trained on synthetic data may achieve high accuracy while depending on spurious correlations or altered feature importance. Explainable AI (XAI) methods have been proposed to understand and audit model behavior. Ribeiro et al. introduced LIME as a model-agnostic local explainer \cite{ribeiro2016lime}, and Lundberg and Lee proposed SHapley Additive exPlanations (SHAP), which is a unified framework grounded in cooperative game theory \cite{lundberg2017shap}. Carvalho et al. conducted a comprehensive survey of interpretability methods and underscored their relevance for trustworthy machine learning \cite{carvalho2019interpretability}.

Building on these insights, this study introduces the SHAP Distance as a semantic fidelity metric for synthetic tabular data. Specifically, the SHAP Distance measures the divergence between global attribution profiles derived from models trained on real versus synthetic datasets. By leveraging established attribution techniques \cite{lundberg2017shap,ribeiro2016lime}, our metric directly evaluates whether downstream models preserve the same reasoning patterns as in those original data. Through empirical validation on enterprise, healthcare, and telecom datasets, we show that SHAP Distance reveals semantic gaps that are invisible to traditional metrics, thereby offering actionable guidance for the development of more faithful synthetic data generators \cite{xu2019ctgan,choi2017ehr,kotelnikov2023tabddpm}.

\section{Related Work}
Research on synthetic tabular data spans three key directions: generative models, evaluation protocols, and explainable AI (XAI) approaches for assessing fidelity and usability.

\subsection{Generative Models}  
A wide variety of methods have been proposed for synthesizing tabular data. Classical approaches include Bayesian networks with privacy guarantees such as PrivBayes \cite{zhang2017privbayes}, and differential privacy based GANs such as PATE-GAN \cite{jordon2019pategan}. VAEs have been extended to tabular data (TVAE) \cite{kiran2023ganvae}, whereas DoppelGANger focuses on high-fidelity temporal tabular synthesis \cite{lin2020doppelganger}. TableGAN includes convolutional architectures to capture local dependencies \cite{park2018tablegan}, whereas Gaussian Copula models have been introduced into the SDV framework \cite{patki2016sdv}. Language models have been adapted for few-shot tabular learning \cite{brown2020gpt3}, and large models have shown promising results in generating synthetic structured data \cite{miletic2024llm}. More recently, TabPFN was proposed as a foundation model for small-data tabular tasks, offering benchmarks for both predictive and generative capacities \cite{hollmann2025tabpfn}.

\subsection{Benchmarks and Evaluation}  
The evaluation of the synthetic data quality has motivated multiple benchmarks. Sidorenko et al. introduce a multi-dimensional evaluation framework that includes distributional similarity, embeddings, and nearest-neighbor metrics \cite{sidorenko2025benchmarking}. Synthcity compares generative models across diverse use cases \cite{qian2023synthcity}, and Hansen et al. demonstrate the role of data-centric profiling in shaping fidelity \cite{hansen2023reimagining}. TabSynDex \cite{chundawat2022tabsyndex} provides a universal score for tabular data evaluation, and a correlation/mean-aware loss has been proposed to enhance the GAN-based synthesis \cite{vu2024corrmean}. In healthcare, Hernandez et al. emphasized domain-specific evaluation challenges \cite{hernandez2023evaluation}, and SynthRO provides a dashboard for ranking synthetic health datasets \cite{santangelo2025synthro}. Vallevik comprehensively reviewed the quality evaluation metrics for synthetic healthcare data \cite{vallevik2024review}.

\subsection{Explainability and Trust}  
XAI methods offer a semantic lens for evaluation. Integrated Gradients \cite{sundararajan2017ig}, DeepLIFT \cite{shrikumar2017deeplift}, Anchors \cite{ribeiro2018anchors}, and counterfactuals \cite{wachter2017counterfactual} have been widely applied to tabular models. Adebayo et al. stress the need for sanity checks of saliency maps \cite{adebayo2018sanity}, whereas Rudin argued for interpretable-by-design models \cite{rudin2019stop}. O’Brien Quinn et al. surveyed explainable analysis methods for tabular data \cite{obrien2024xai}, and Velmurugan et al. proposed guidelines for the grounded evaluation of XAI techniques \cite{velmurugan2024guidelines}. Comparative studies have demonstrated the variability in attribution across XAI methods in tabular healthcare  tasks \cite{qureshi2025explainability}, reinforcing the need for semantic fidelity measures that focus on reasoning consistency rather than statistical similarity alone. Tu et al. further showed that causal benchmarks expose structural gaps in synthetic data \cite{tu2024caustab}.

Together, these studies motivate our contribution: introducing the SHAP Distance as a semantic fidelity metric for synthetic tabular data that complements distributional and utility-based evaluations.

\section{Methodology}
In this section, we present our proposed methodology for evaluating the semantic fidelity of synthetic tabular data. The pipeline comprises three main components: data preprocessing, attribution-based analysis, and SHAP Distance computation. Intuitively, the SHAP Distance quantifies the average absolute difference in normalized SHAP attributions across corresponding features, providing a semantic measure of reasoning alignment beyond distributional resemblance.

\subsection{Problem Formulation}
Let $D_{\text{real}} = \{ (x_i, y_i) \}_{i=1}^{N}$ denote the real dataset with $N$ samples, where $x_i \in \mathbb{R}^d$ is a $d$-dimensional feature vector and $y_i$ is the target label. A synthetic dataset $D_{\text{syn}} = \{ (x_j', y_j') \}_{j=1}^{M}$ is generated using a synthetic data generator.  

The objective is to evaluate the semantic alignment between $D_{\text{real}}$ and $D_{\text{syn}}$. Traditional statistical measures (e.g., marginal distributions and correlation matrices) fail to capture whether predictive logic is preserved. We propose the SHAP Distance, a measure based on feature attribution consistency.

\subsection{Attribution-Based Evaluation}
To compare the reasoning between real and synthetic data, we train two classifiers:
\[
\mathcal{C}_{\text{real}} \leftarrow \text{Train}(D_{\text{real}}), 
\quad 
\mathcal{C}_{\text{syn}} \leftarrow \text{Train}(D_{\text{syn}}).
\]

For each classifier, the SHAP values are computed to quantify the contribution of each feature to the prediction. Let $\phi^{\text{real}}_k$ denote the average SHAP value of feature $k$ in $\mathcal{C}_{\text{real}}$, and $\phi^{\text{syn}}_k$ denote the same for $\mathcal{C}_{\text{syn}}$.

This yields two attribution vectors:
\[
\boldsymbol{\phi}_{\text{real}} = [\phi^{\text{real}}_1, \ldots, \phi^{\text{real}}_d],
\quad
\boldsymbol{\phi}_{\text{syn}} = [\phi^{\text{syn}}_1, \ldots, \phi^{\text{syn}}_d].
\]

\subsection{SHAP Distance Metric}
We quantify the semantic alignment by computing the cosine distance between attribution vectors:
\[
  D_{\mathrm{SHAP}} = 1 - \frac{\boldsymbol{\phi}_{\mathrm{real}} \cdot \boldsymbol{\phi}_{\mathrm{syn}}}
                         {\|\boldsymbol{\phi}_{\mathrm{real}}\|\;\|\boldsymbol{\phi}_{\mathrm{syn}}\|},
\]
where $\boldsymbol{\phi}_{\mathrm{real}}$ and $\boldsymbol{\phi}_{\mathrm{syn}}$ are the SHAP attribution vectors obtained from the classifiers trained on the real and synthetic datasets, respectively.  

The metric $D_{\mathrm{SHAP}}$ measures the angular dissimilarity between the two attribution vectors. Values closer to 0 indicate that the synthetic data preserves the attribution pattern of the real data, thereby maintaining semantic fidelity. Conversely, values closer to 1 suggest that the decision logic derived from synthetic data diverges significantly from that of real data, even if the statistical distributions appear similar.

\subsection{Iterative Refinement Framework}
Although $D_{\text{SHAP}}$ can be used as a standalone metric, we extend it into a feedback loop to improve the synthetic generation. The procedure is as follows:

\begin{enumerate}
    \item Generate an initial synthetic dataset $D_{\text{syn}}^{(0)}$.
    \item Train classifiers $\mathcal{C}_{\text{real}}$ and $\mathcal{C}_{\text{syn}}^{(t)}$.
    \item Compute SHAP Distance $D_{\text{SHAP}}^{(t)}$.
    \item If $D_{\text{SHAP}}^{(t)} > \epsilon$, refine the generation process by emphasizing divergent features in the prompt or model configuration.
    \item Repeat until $D_{\text{SHAP}}^{(t)} \leq \epsilon$ or a maximum number of iterations $T$ is reached.
\end{enumerate}

This iterative process aligns the feature attribution patterns between the real and synthetic data, producing semantically faithful synthetic datasets.

\section{Experiments}
We conducted experiments to validate the effectiveness of the SHAP Distance as a semantic fidelity metric for synthetic tabular data. Our study investigates the following three research questions:  
(1) Does the SHAP Distance correlate with model performance gaps between real and synthetic data?  
(2) Can the SHAP Distance capture semantic inconsistencies that statistical measures fail to identify?  
(3) Is the SHAP Distance stable across datasets with different structures and domains?

\subsection{Datasets}
The proposed method was evaluated using three representative tabular datasets:  

\begin{itemize}
    \item \textbf{UCI Heart Disease:} This dataset originates from the Cleveland Clinic Foundation and includes biomedical records for 303 patients. Each record describes attributes such as age, cholesterol level, resting blood pressure, and electrocardiogram results. The binary target indicates whether the patient has a heart condition. Owing to its small size and imbalance, this dataset is well-suited for stress-testing generative methods in medically sensitive, low-resource scenarios.
    \item \textbf{Enterprise Invoice Usage:} This proprietary dataset consists of behavioral records of 500 companies using an enterprise SaaS invoicing platform in Japan. Key features include the monthly invoice activity (\texttt{InvoiceCount\_1Month}), total six-month invoice usage (\texttt{InvoiceCount\_6MonthTotal}), the number of total users, IDs, and partner types, as well as free and paid user counts. The binary target variable \texttt{InvoiceIssued} denotes whether the company has issued any paid invoices. This dataset reflects real-world enterprise behavior and is suitable for evaluating semantic consistency in structured data synthesis.
    \item \textbf{Telco Churn:}  This open dataset provides subscription and billing details for over 7,000 customers from a telecom service provider. It includes both categorical variables (e.g., \texttt{Contract}, \texttt{InternetService}) and numerical values (e.g., \texttt{MonthlyCharges}). The prediction task involves determining whether a customer will churn. The diversity and volume of features make it a suitable benchmark for evaluating the robustness of synthetic generation under mixed data types.

\end{itemize}

All datasets were preprocessed using missing-value imputation, categorical encoding, and normalization. Class imbalance was addressed by random undersampling.

\subsection{Synthetic Data Generation}
For synthetic data generation, we adopted our previously proposed framework, KGSynX~\cite{yu2025kgsynx}. 
This framework integrates knowledge graph guidance and attribution-based refinement, and was empirically validated in our earlier work to achieve superior downstream utility compared with baseline generative models. 

KGSynX follows a four-step pipeline:

(Step 1) \textit{Knowledge Graph Construction}: Raw tables are lifted into a knowledge graph where each row becomes an entity node and each attribute-value is linked via typed relations, thereby exposing domain rules and constraints. 

(Step 2) \textit{Embedding \& Initial Synthesis}: Node2Vec embeddings are computed over the knowledge graph and injected into the prompts for ChatGPT-4o to generate an initial batch of synthetic records. 

(Step 3) \textit{SHAP Analysis \& Prompt Feedback Loop}: classifiers are trained on real and synthetic data, and SHAP values are used to measure feature-importance gaps, which are automatically translated into targeted prompt edits, and Steps 2–3 are repeated the until misalignment falls below a preset threshold. 

(Step 4) \textit{Final Optimized Synthetic Data Generation}: Using the refined prompts, the final synthetic dataset is produced, matching the real data in both statistical properties and decision-logic semantics. 
This modular design decouples knowledge extraction, generation, feedback, and refinement while preserving end-to-end semantic guidance. 

By leveraging this structured pipeline, we ensure that the evaluation of the SHAP Distance in this study was conducted on high-quality synthetic datasets.

\subsection{Evaluation Metrics}
We assessed both the statistical and semantic fidelity using the following three metrics:  

\begin{itemize}
    \item \textbf{Kullback-Leibler (KL) Divergence:} Measures the marginal distributional similarity.  
     \item \textbf{Principal Component Analysis (PCA) Variance Ratio:} Captures the global geometric structure by comparing the variance explained by the first two principal components of the real and synthetic data.
    \item \textbf{SHAP Distance:} Measures semantic fidelity based on feature attribution alignment.  
\end{itemize}

\section{Results}

\subsection{Marginal Distributions and KL Divergence}

\begin{figure}[H]
    \centering
    \includegraphics[width=0.95\linewidth]{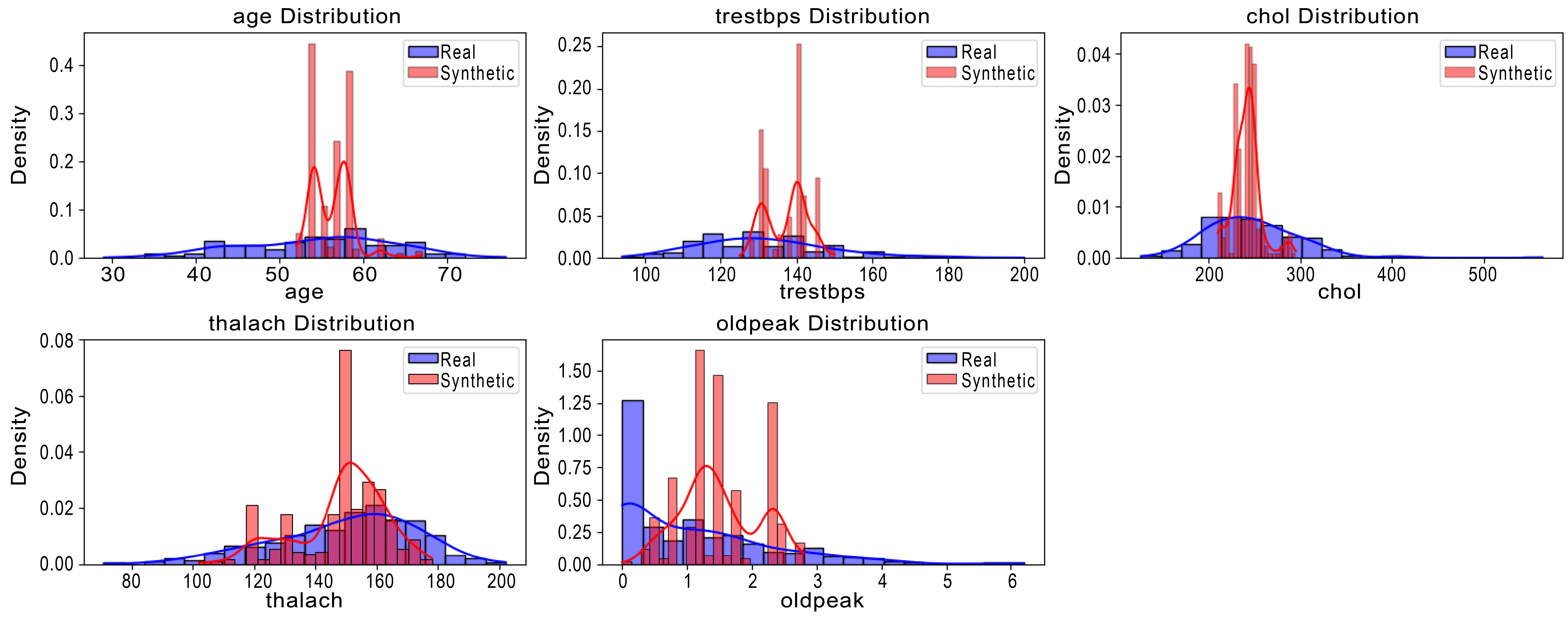}
    \caption{Density comparison (UCI Heart Disease).}
    \label{fig:uci_marginal}
\end{figure}

\begin{figure}[H]
  \centering
  \includegraphics[width=0.95\linewidth]{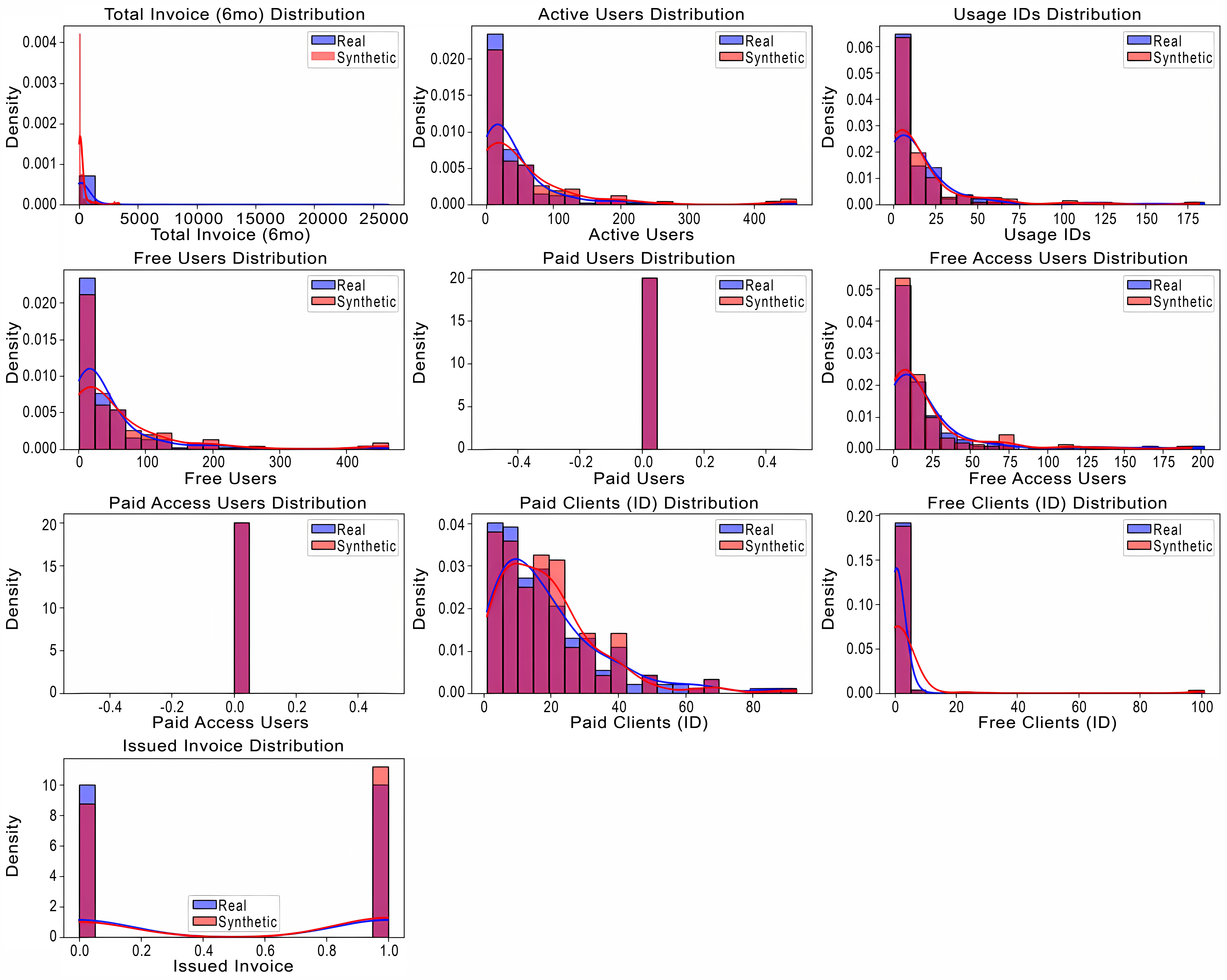}
  \caption{Density comparison (Enterprise Invoice).}
  \label{fig:inf_marginal}
\end{figure}

\begin{figure}[H]
  \centering
  \includegraphics[width=0.95\linewidth]{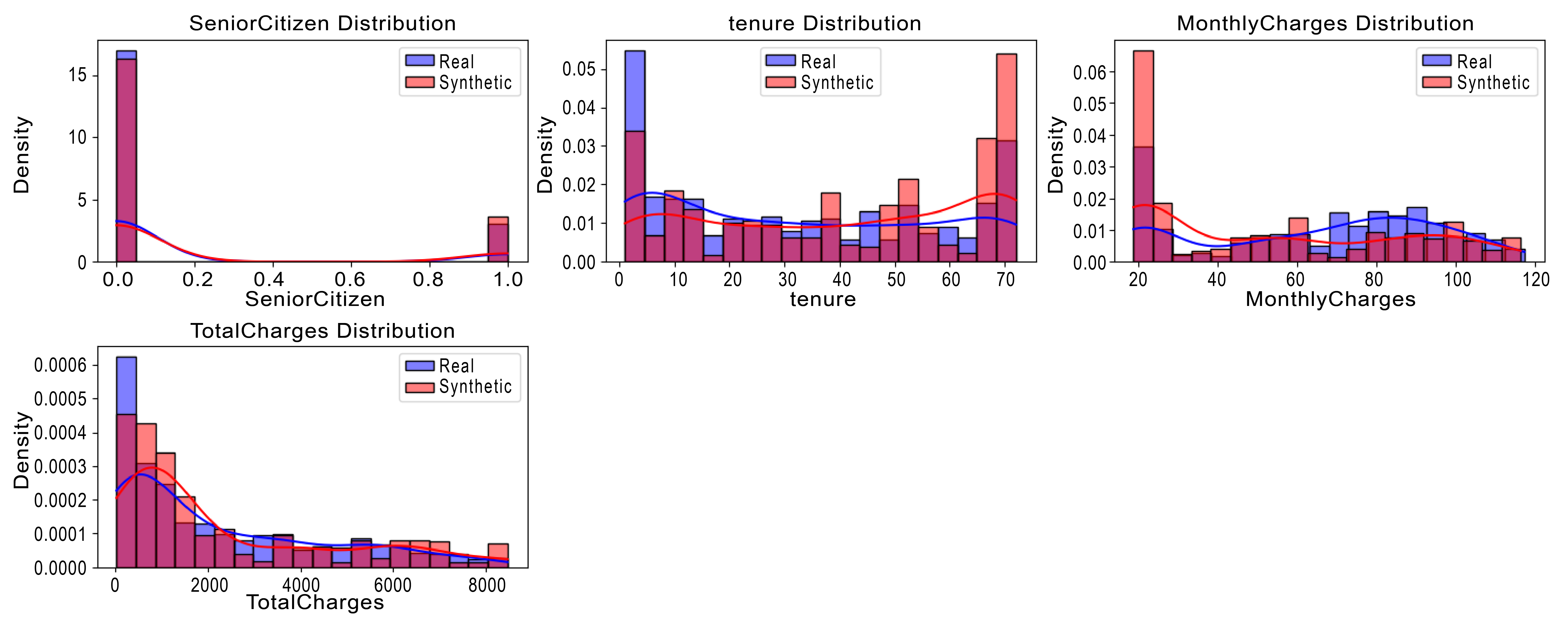}
  \caption{Density comparison (Telco Churn).}
  \label{fig:telco_marginal}
\end{figure}

We begin by analyzing the marginal distributions of key numeric features across the three datasets. Figures~\ref{fig:uci_marginal}, \ref{fig:inf_marginal}, and \ref{fig:telco_marginal} show density plots comparing the real and synthetic distributions. Overall, the synthetic data captures the central tendencies well but often exhibited reduced variability in the tails, leading to smoother peaks and  an underrepresentation of extreme values.

To quantify the alignment, we computed the KL divergence for the selected major features from each dataset. Table~\ref{tab:kl_all} summarizes the scores.  

\begin{table}[H]
\centering
\caption{KL divergence across datasets}
\label{tab:kl_all}
\begin{tabular}{lcc}
\toprule
\textbf{Dataset} & \textbf{Feature} & \textbf{KL Divergence} \\
\midrule
\multirow{5}{*}{UCI Heart Disease} 
 & Age            & 5.3840 \\
 & Resting BP     & 6.0128 \\
 & Cholesterol    & 0.9035 \\
 & Max Heart Rate & 0.4967 \\
 & ST Depression  & 4.6848 \\
\midrule
\multirow{6}{*}{Enterprise Invoice Usage} 
 & Total Invoices (6mo) & 0.0519 \\
 & Active Users         & 0.1698 \\
 & Usage IDs            & 0.2387 \\
 & Free Users           & 0.1698 \\
 & Paid Clients (ID)    & 0.4072 \\
 & Issued Invoice (bin) & 0.0073 \\
\midrule
\multirow{4}{*}{Telco Churn} 
 & SeniorCitizen   & 0.0032 \\
 & Tenure          & 0.1616 \\
 & MonthlyCharges  & 0.2991 \\
 & TotalCharges    & 0.1227 \\
\bottomrule
\end{tabular}
\end{table}

The results revealed distinct patterns. In the UCI Heart Disease dataset, features such as \texttt{blood pressure} and \texttt{age} exhibited high KL values, reflecting distributional mismatches in tail regions. In the Enterprise Invoice Usage dataset, most KL scores are below 0.2, with the exception of the \texttt{Paid Client} identifiers, suggesting stronger deviations in rare-value distributions. The Telco Churn dataset showed consistently low KL scores, with the lowest in the binary \texttt{SeniorCitizen} feature, and slightly higher values in long-tailed variables such as \texttt{MonthlyCharges}.  

These findings highlight that although marginal statistics are generally well preserved, specific high-variance or long-tailed features remain challenging for synthetic data generation.

\subsection{PCA Projection of Global Geometry}

\begin{figure*}[htbp]
  \centering
  \begin{subfigure}{0.32\linewidth}
    \centering
    \includegraphics[width=\linewidth]{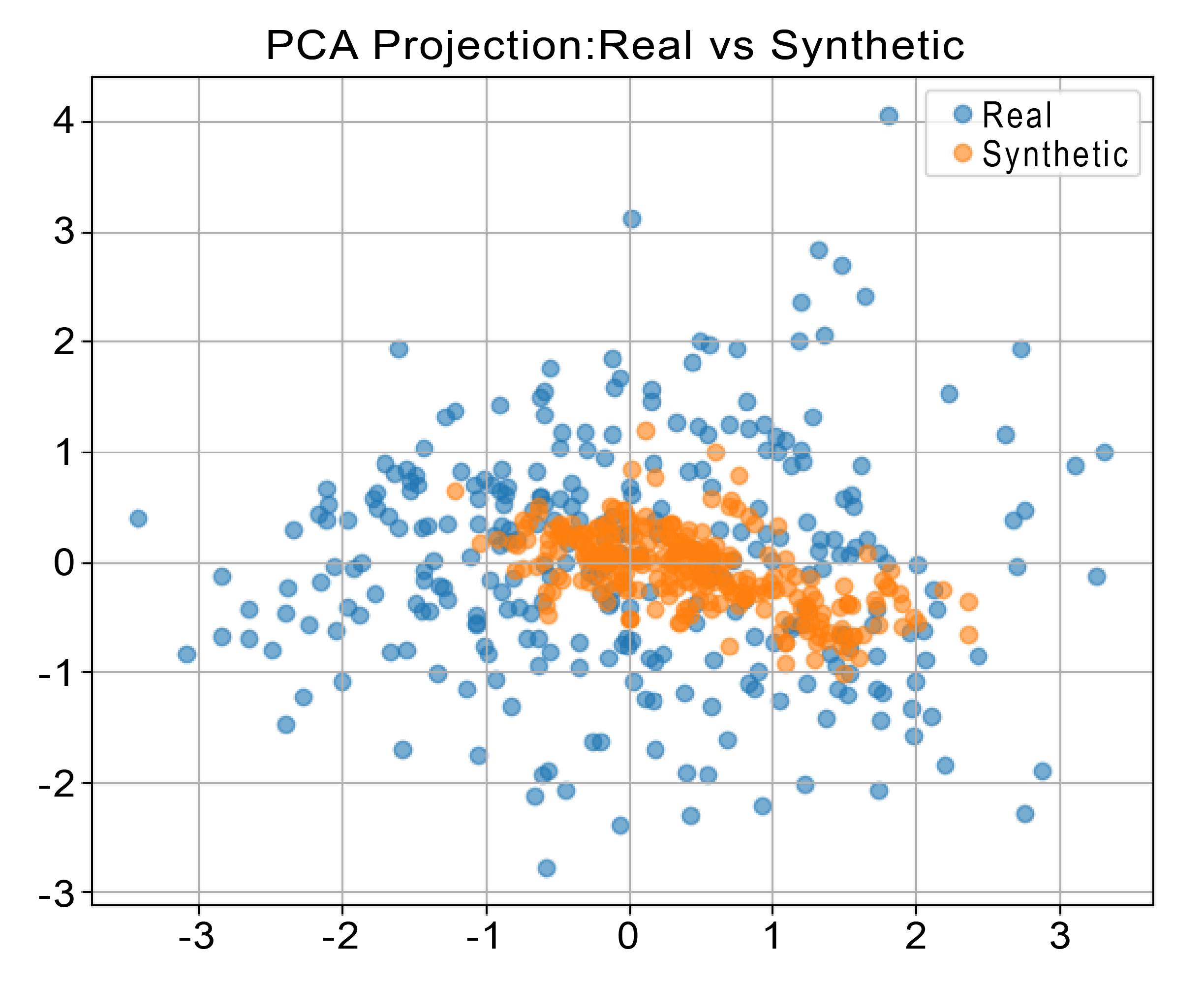}
    \caption{UCI Heart Disease.}
    \label{fig:uci_pca}
  \end{subfigure}
  \hfill
  \begin{subfigure}{0.32\linewidth}
    \centering
    \includegraphics[width=\linewidth]{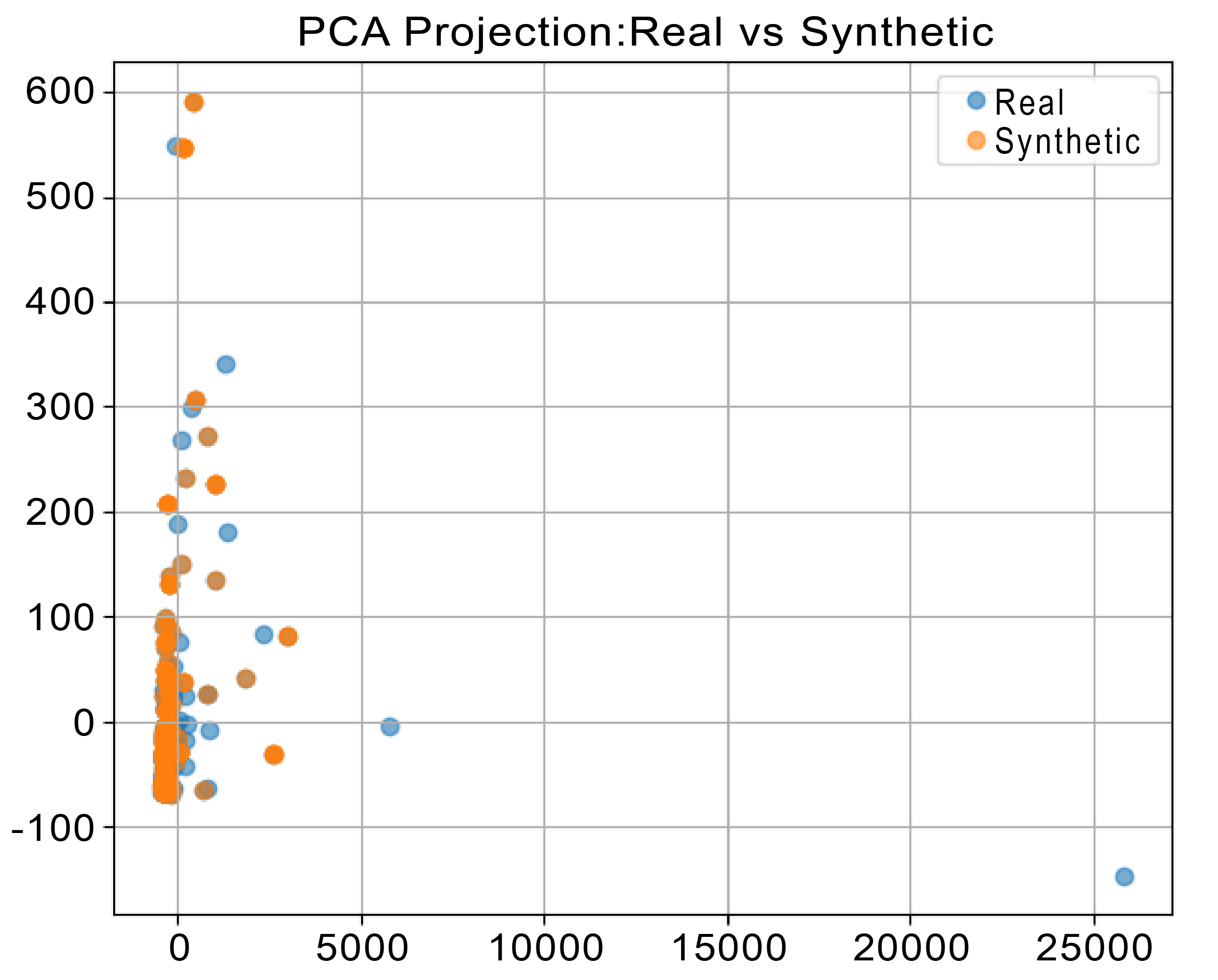}
    \caption{Enterprise Invoice.}
    \label{fig:inf_pca}
  \end{subfigure}
  \hfill
  \begin{subfigure}{0.32\linewidth}
    \centering
    \includegraphics[width=\linewidth]{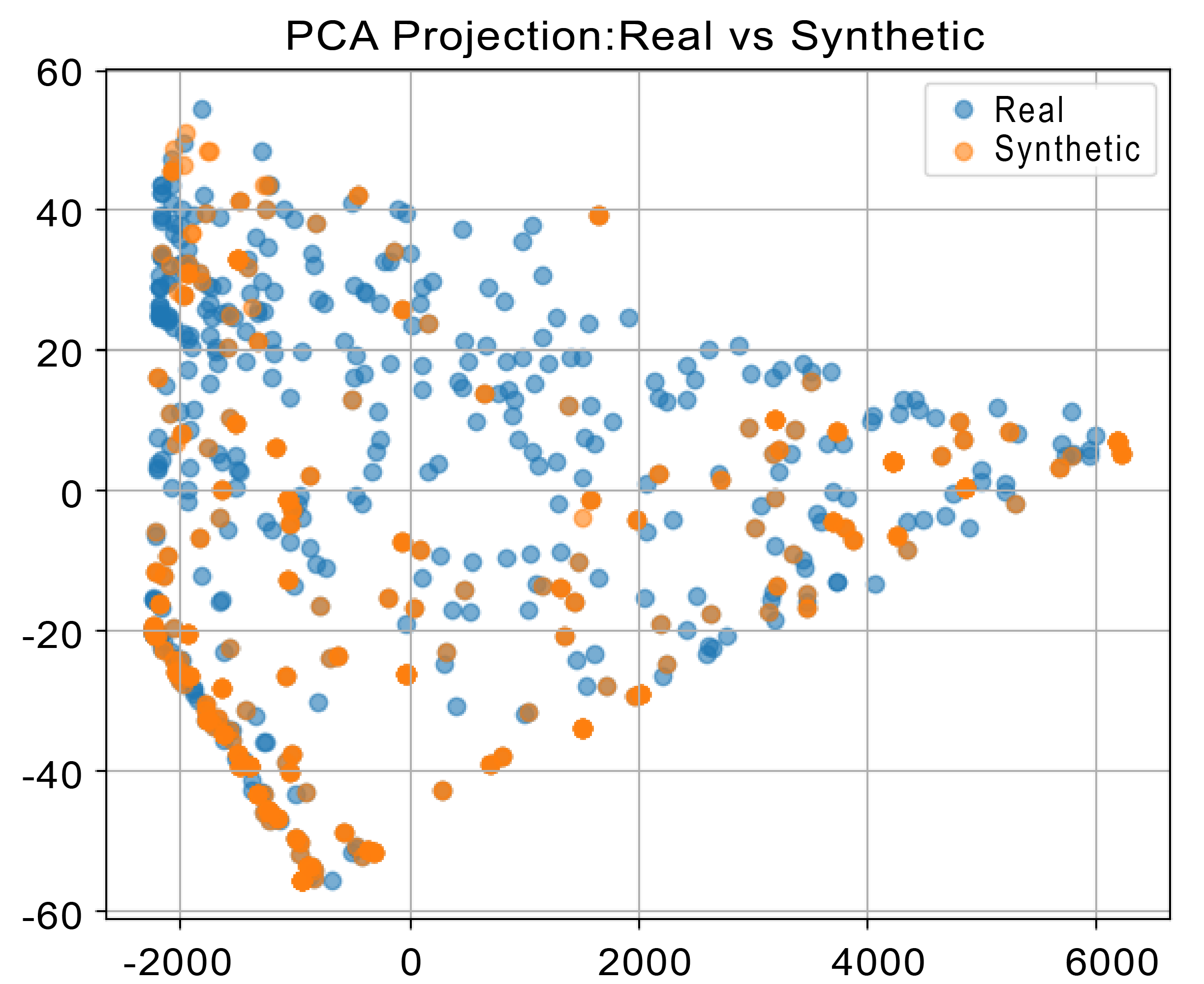}
    \caption{Telco Churn.}
    \label{fig:telco_pca}
  \end{subfigure}

  \caption{PCA projections of real and synthetic data across three domains: 
  (a) healthcare (UCI Heart Disease), 
  (b) enterprise (Invoice Usage), 
  and (c) telecommunications (Telco Churn).}
  \label{fig:pca_overview}
\end{figure*}

To investigate the global geometric structure of the data further, we applied PCA to both the real and synthetic samples and projected them onto a two-dimensional space. Figures~\ref{fig:uci_pca}, \ref{fig:inf_pca}, and \ref{fig:telco_pca} show the resulting projections for the three datasets.  

\begin{figure*}[htbp]
  \centering
  \begin{subfigure}{0.32\linewidth}
    \centering
    \includegraphics[width=\linewidth]{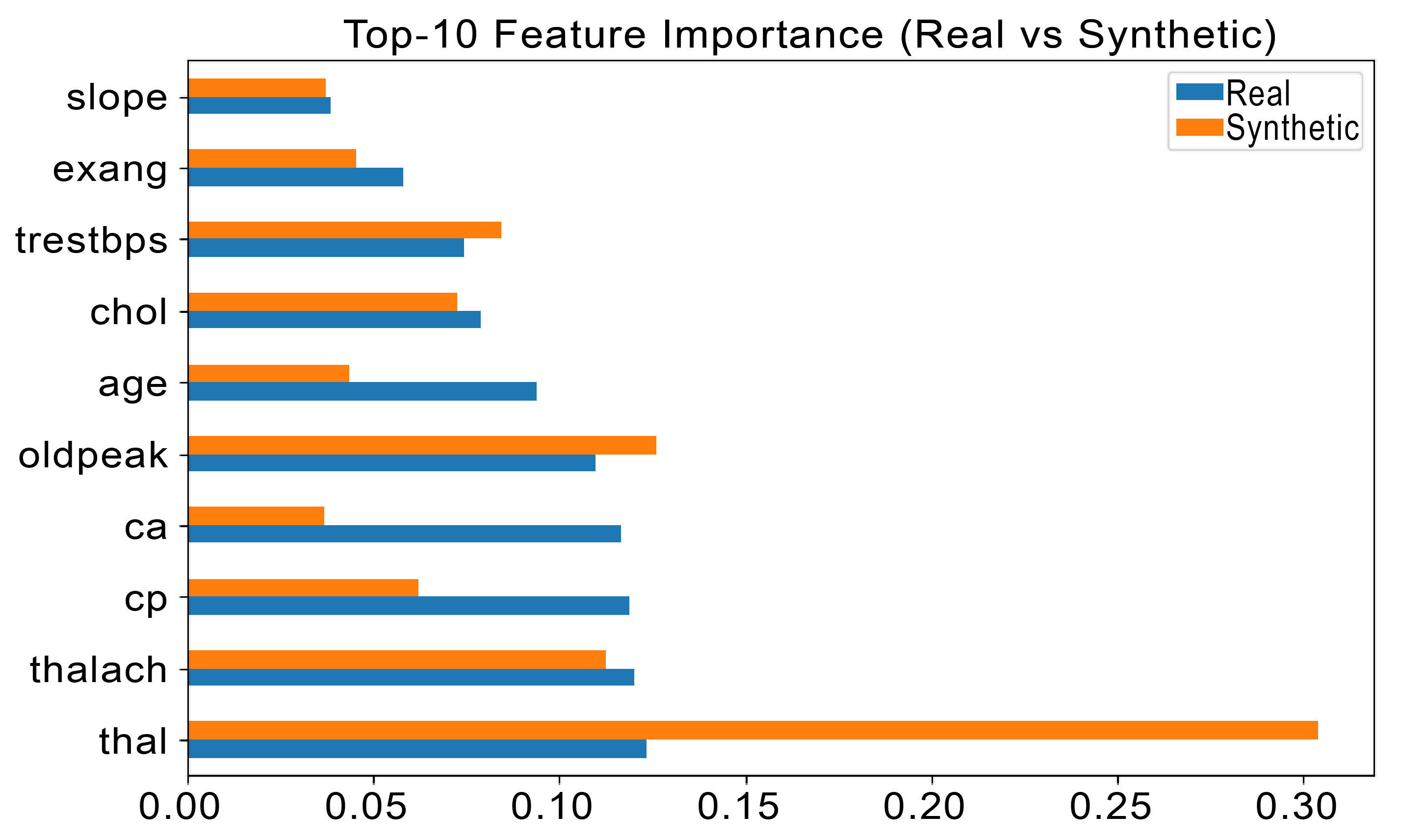}
    \caption{UCI Heart Disease.}
    \label{fig:uci_shap}
  \end{subfigure}
  \hfill
  \begin{subfigure}{0.32\linewidth}
    \centering
    \includegraphics[width=\linewidth]{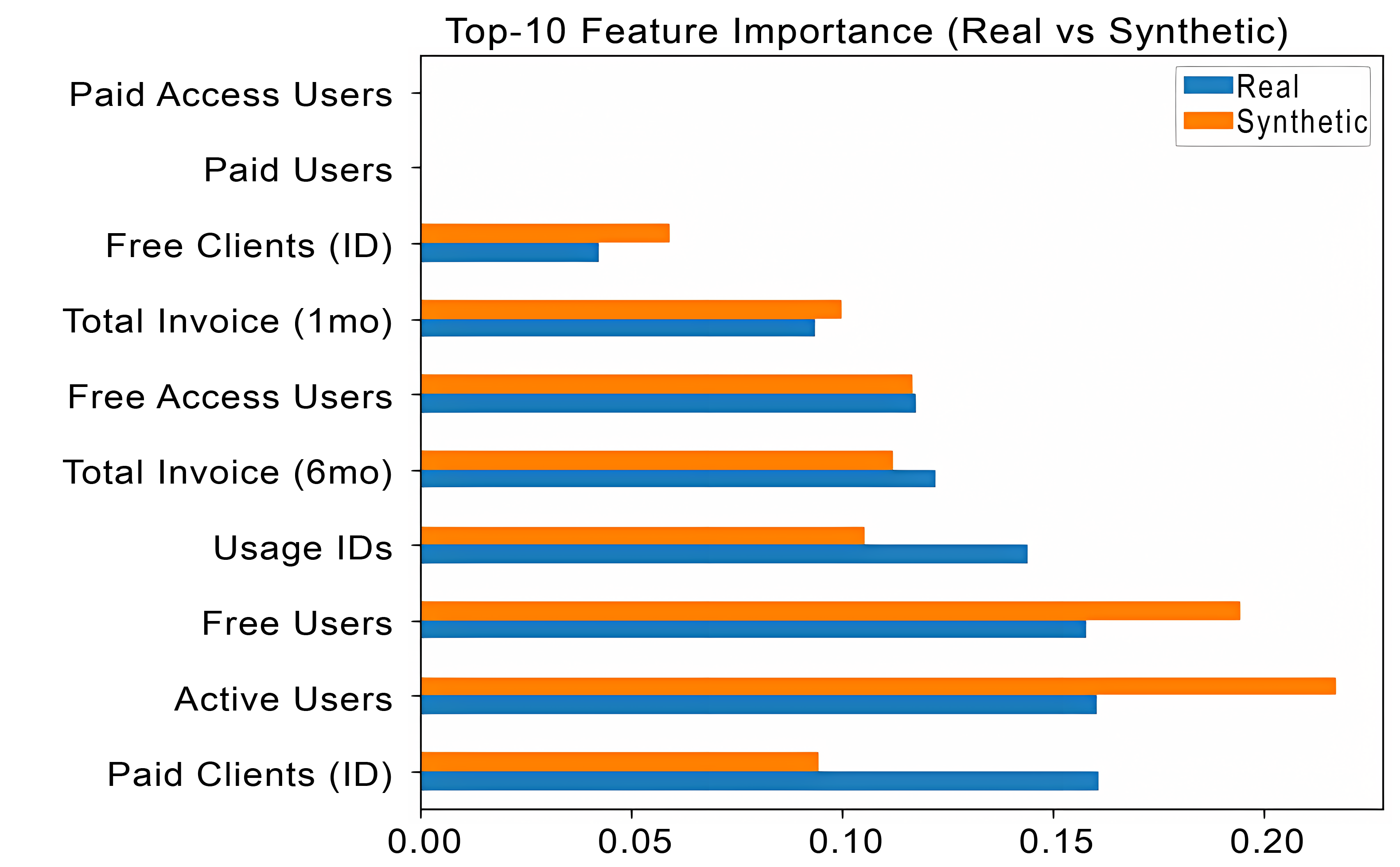}
    \caption{Enterprise Invoice.}
    \label{fig:inf_shap}
  \end{subfigure}
  \hfill
  \begin{subfigure}{0.32\linewidth}
    \centering
    \includegraphics[width=\linewidth]{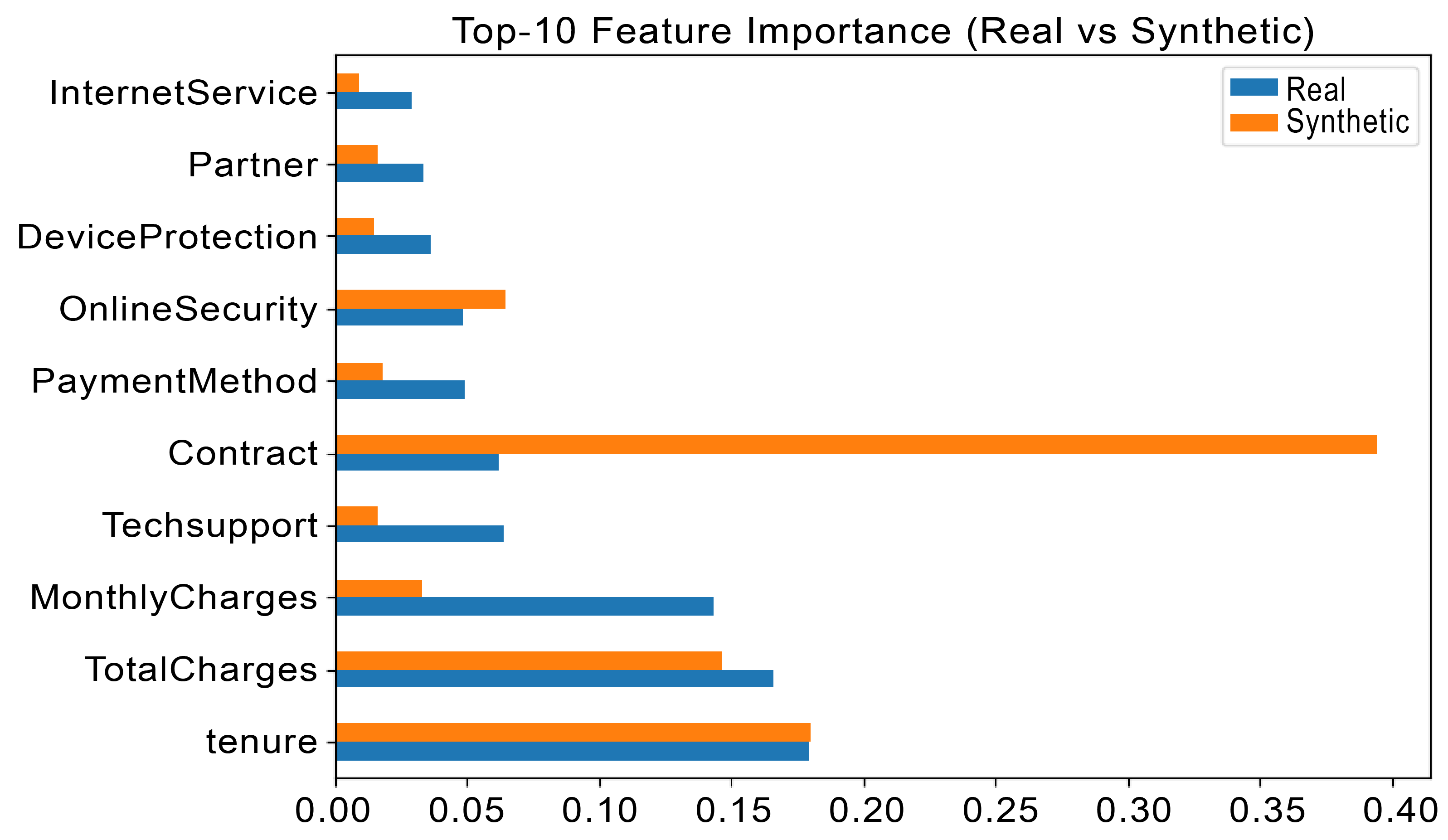}
    \caption{Telco Churn.}
    \label{fig:telco_shap}
  \end{subfigure}

  \caption{SHAP feature attribution comparison across three datasets: 
  (a) healthcare (UCI Heart Disease), 
  (b) enterprise (Invoice Usage), 
  and (c) telecommunications (Telco Churn).}
  \label{fig:shap_overview}
\end{figure*}

Across datasets, real data consistently shows a broader spread, capturing multiple directions of variability, while synthetic data exhibits tighter clustering, indicating concentration around dominant variance directions. This behavior highlights a common limitation of synthetic generators, namely the underrepresentation of rare patterns or peripheral regions.  

To quantify these observations, Table~\ref{tab:pca_all} reports the explained variance ratios of the first two principal components for the real and synthetic data.  

\begin{table}[htbp]
\centering
\caption{Explained variance ratio of the first two principal components across datasets}
\label{tab:pca_all}
\begin{tabular}{llccc}
\toprule
\textbf{Dataset} & \textbf{Type} & \textbf{PC1 (\%)} & \textbf{PC2 (\%)} & \textbf{PC1+PC2} \\
\midrule
\multirow{2}{*}{UCI Heart Disease} 
 & Real      & 35.94 & 21.94 & 57.88 \\
 & Synthetic & 64.84 & 13.89 & 78.73 \\
\midrule
\multirow{2}{*}{Enterprise Invoice} 
 & Real      & 99.69 & 0.28  & 99.97 \\
 & Synthetic & 64.84 & 13.89 & 78.73 \\
\midrule
\multirow{2}{*}{Telco Churn} 
 & Real      & 99.99 & 0.01  & 100.00 \\
 & Synthetic & 64.84 & 13.89 & 78.73 \\
\bottomrule
\end{tabular}
\end{table}

These results confirm three patterns. First, in the UCI Heart Disease dataset, the synthetic data collapses much of the variance into the first component, reducing the spread across secondary directions. Second, in the Enterprise Invoice Usage dataset, the extreme dominance of PC1 in the real data were not preserved in the synthetic data, suggesting the undercoverage of rare usage behaviors. Third, in the Telco Churn dataset, the real data were almost one-dimensional, and while the synthetic data followed the main axis, they introduced more variance along the secondary components.  

Overall, the PCA analysis demonstrates that the synthetic data tended to concentrate the variance along dominant dimensions, leading to smoother but less diverse representations. This finding reinforces the need for mechanisms that preserve the full covariance structure when generating synthetic tabular data.

\begin{figure*}[htbp]
  \centering
  \begin{subfigure}{0.45\linewidth}
    \centering
    \includegraphics[width=\linewidth]{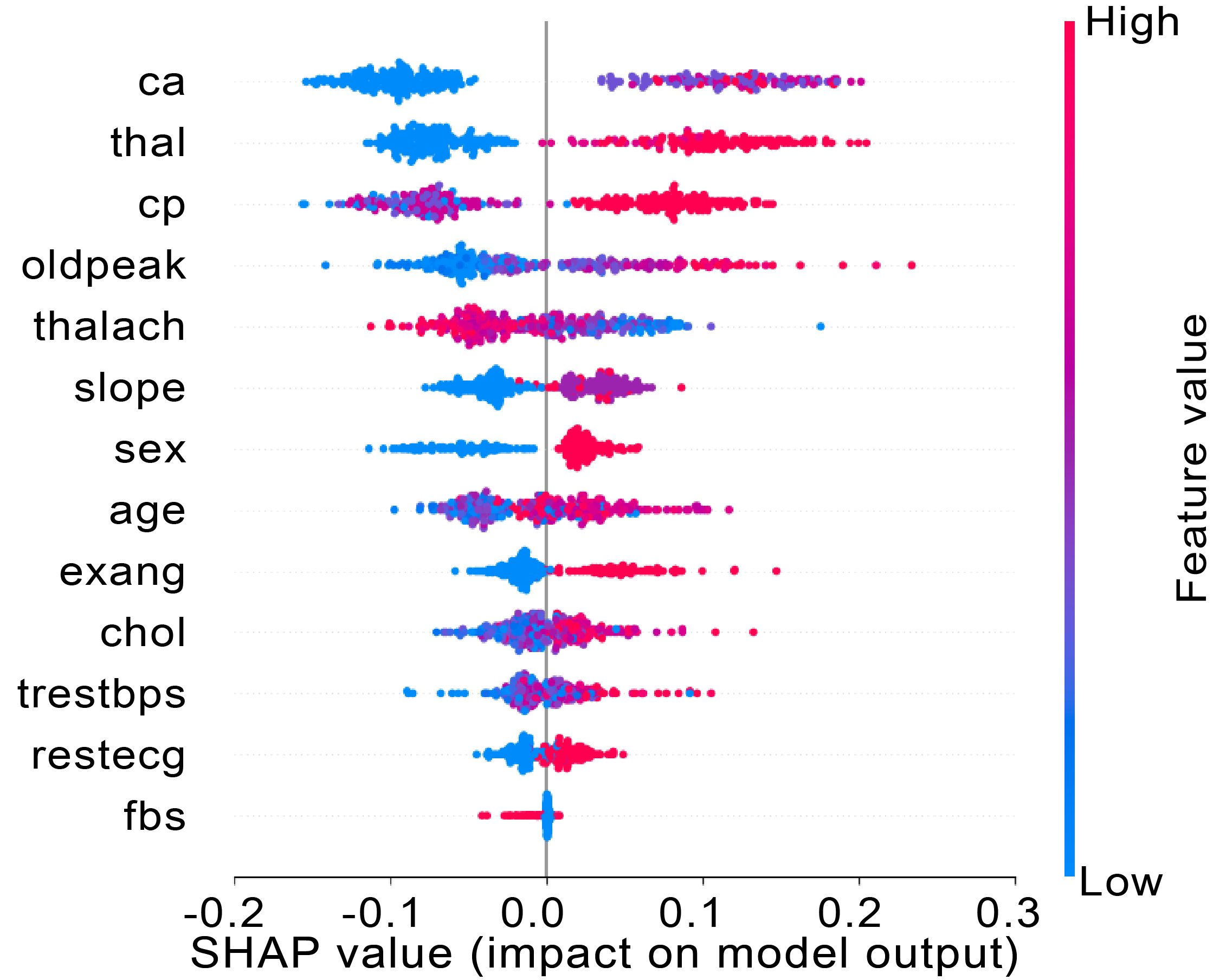}
    \caption{SHAP summary plot for real data (UCI Heart Disease).}
    \label{fig:uci_shap_real_summary}
  \end{subfigure}
  \hfill
  \begin{subfigure}{0.45\linewidth}
    \centering
    \includegraphics[width=\linewidth]{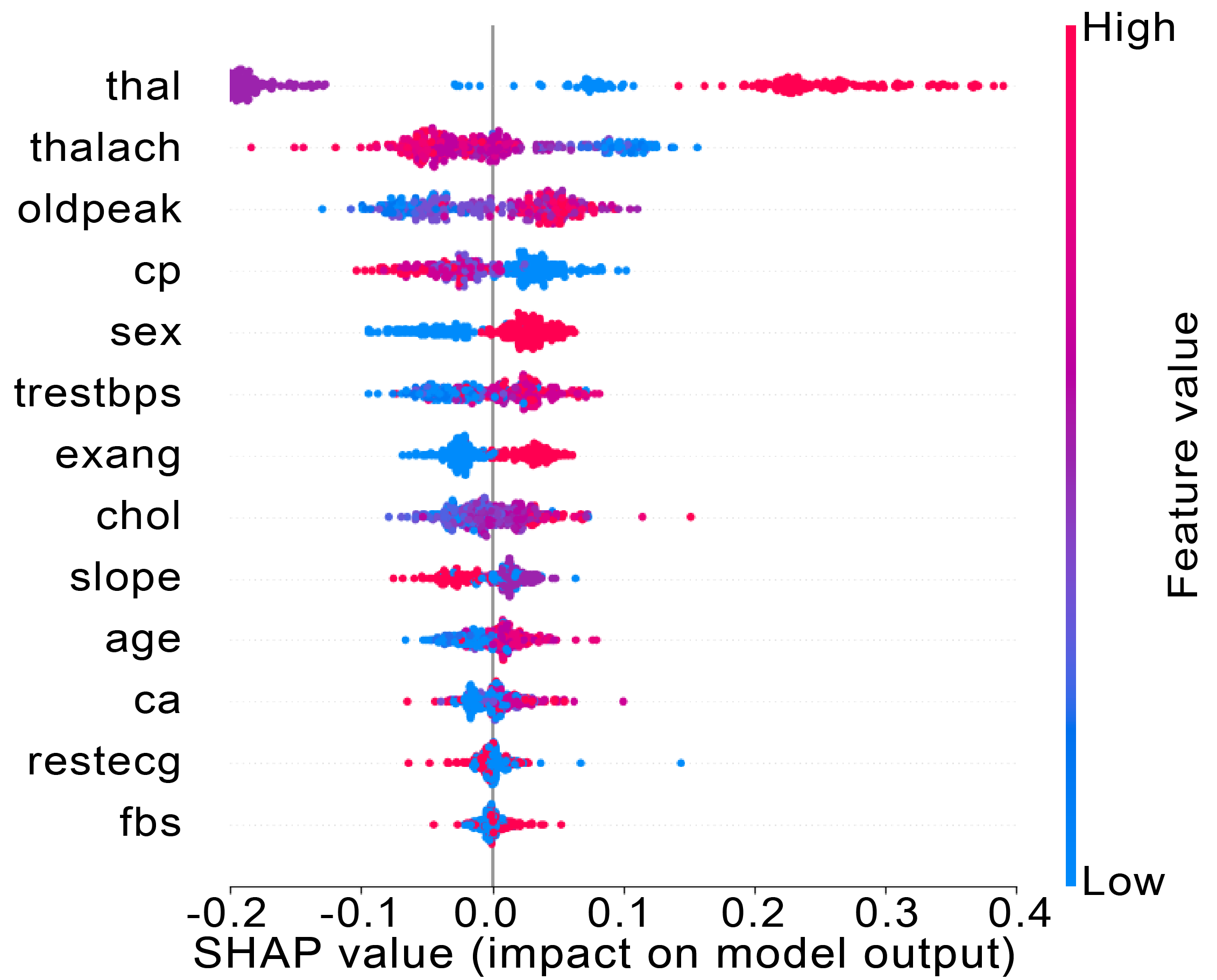}
    \caption{SHAP summary plot for synthetic data (UCI Heart Disease).}
    \label{fig:uci_shap_syn_summary}
  \end{subfigure}

  \vspace{0.5cm} 

  \begin{subfigure}{0.45\linewidth}
    \centering
    \includegraphics[width=\linewidth]{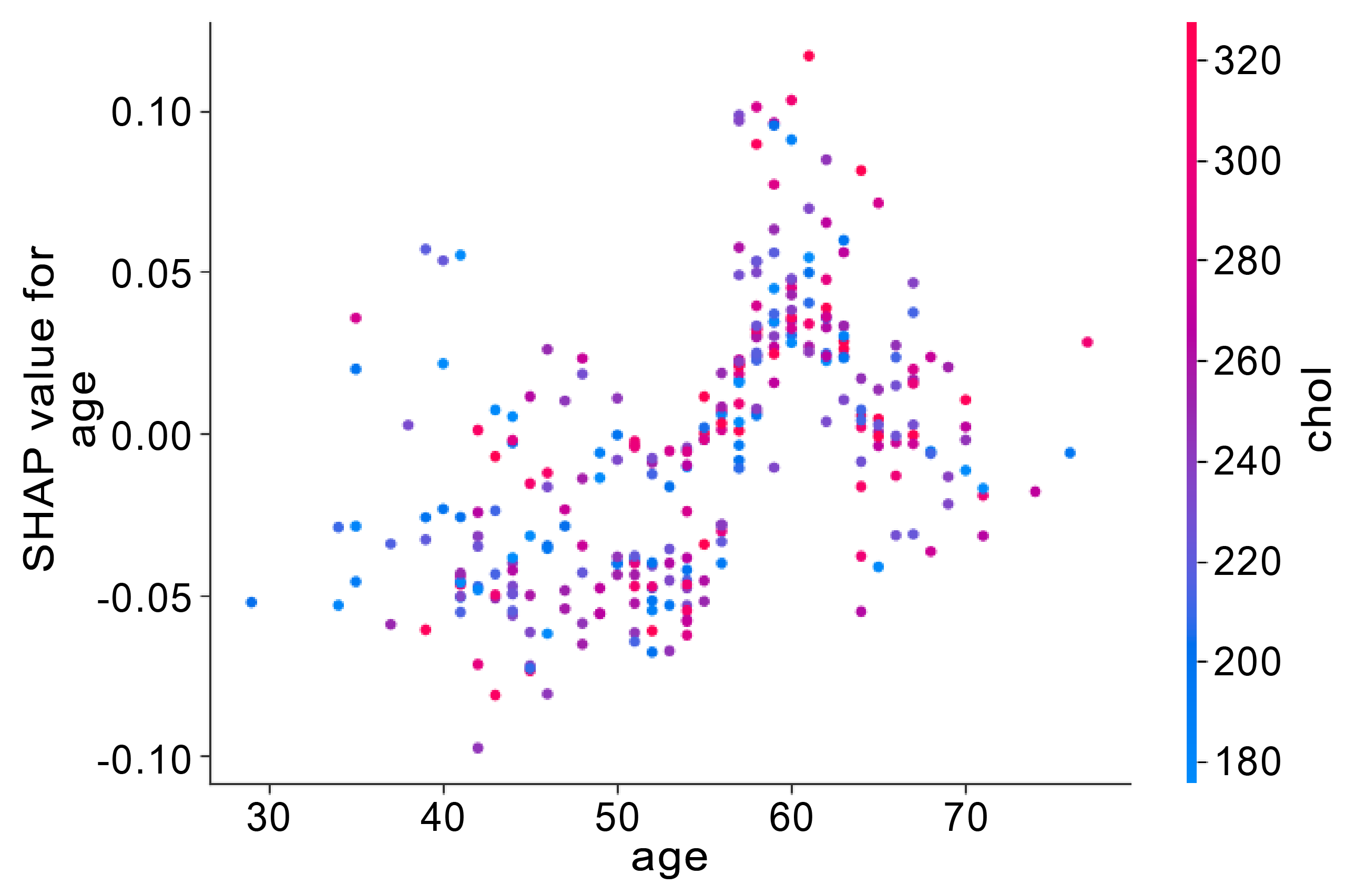}
    \caption{SHAP dependence plot for real data: effect of age (colored by cholesterol).}
    \label{fig:uci_shap_real_dependence}
  \end{subfigure}
  \hfill
  \begin{subfigure}{0.45\linewidth}
    \centering
    \includegraphics[width=\linewidth]{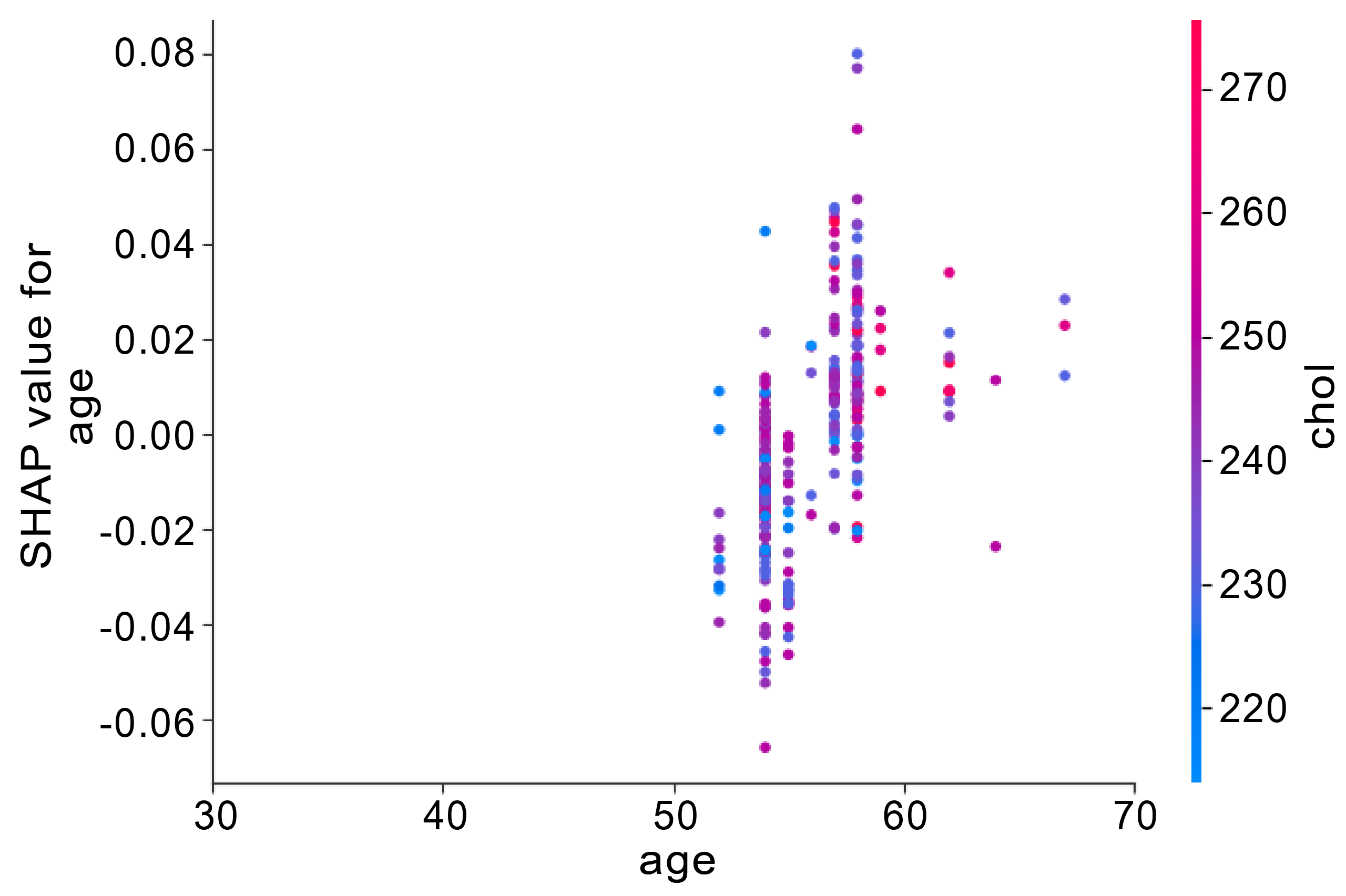}
    \caption{SHAP dependence plot for synthetic data: effect of age (colored by cholesterol).}
    \label{fig:uci_shap_syn_dependence}
  \end{subfigure}

  \caption{SHAP-based feature attribution consistency analysis on UCI Heart Disease dataset: 
  (a) summary plot (real data), 
  (b) summary plot (synthetic data), 
  (c) dependence plot for real data, and 
  (d) dependence plot for synthetic data.}
  \label{fig:uci_shap_overview}
\end{figure*}

\subsection{SHAP Attribution Alignment}

To assess whether classifiers trained on the synthetic data preserved the same decision logic as those trained on real data, we compared the feature attributions using SHAP values. Figures~\ref{fig:uci_shap}, \ref{fig:inf_shap}, and \ref{fig:telco_shap} show the top-10 ranked features across the three datasets.

In the UCI Heart Disease dataset, both the real and synthetic models highlight \texttt{thal}, \texttt{thalach}, and \texttt{cp} as the dominant predictors, indicating strong alignment in the primary decision pathways. However, the synthetic models overemphasize \texttt{oldpeak} and \texttt{trestbps}, suggesting missing correlations in the tail regions.

In the Enterprise Invoice Usage dataset, features such as \texttt{client IDs} and \texttt{usage counts} remained consistently top-ranked across the real and synthetic models. Nevertheless, mid-level deviations appeared in variables such as \texttt{Free Access Users} and the binary \texttt{Invoice Issued}, which may reflect simplified structural prompts.

In the Telco Churn dataset, both models identified \texttt{Contract}, \texttt{tenure}, and \texttt{TotalCharges} as key predictors. However, the synthetic models attribute disproportionately high importance to \texttt{Contract} (almost 40\%), while diminishing the contributions of billing-related features such as \texttt{MonthlyCharges} and \texttt{PaymentMethod}. This indicates a tendency toward prompt-driven bias, though demographic and service-related features (\texttt{Partner}, \texttt{OnlineSecurity}) remained well aligned.

The results revealed that the attribution alignment varies across domains. The UCI Heart Disease dataset exhibited the largest semantic gap, which was primarily driven by continuous variables with clinical dependencies. The Enterprise Invoice dataset showed moderate alignment, with deviations concentrated in less frequent user categories. The Telco Churn dataset achieves the lowest SHAP Distance, confirming that its balanced schema and structured features are more easily preserved by  the synthetic generators.  

Overall, the SHAP-based evaluation provides critical insights into reasoning consistency, which is not captured by marginal or covariance statistics. This highlights the fact that synthetic data preserves predictive pathways faithfully and when prompt biases introduce distortions.

\begin{figure*}[htbp]
  \centering
  \begin{subfigure}{0.45\linewidth}
    \centering
    \includegraphics[width=\linewidth]{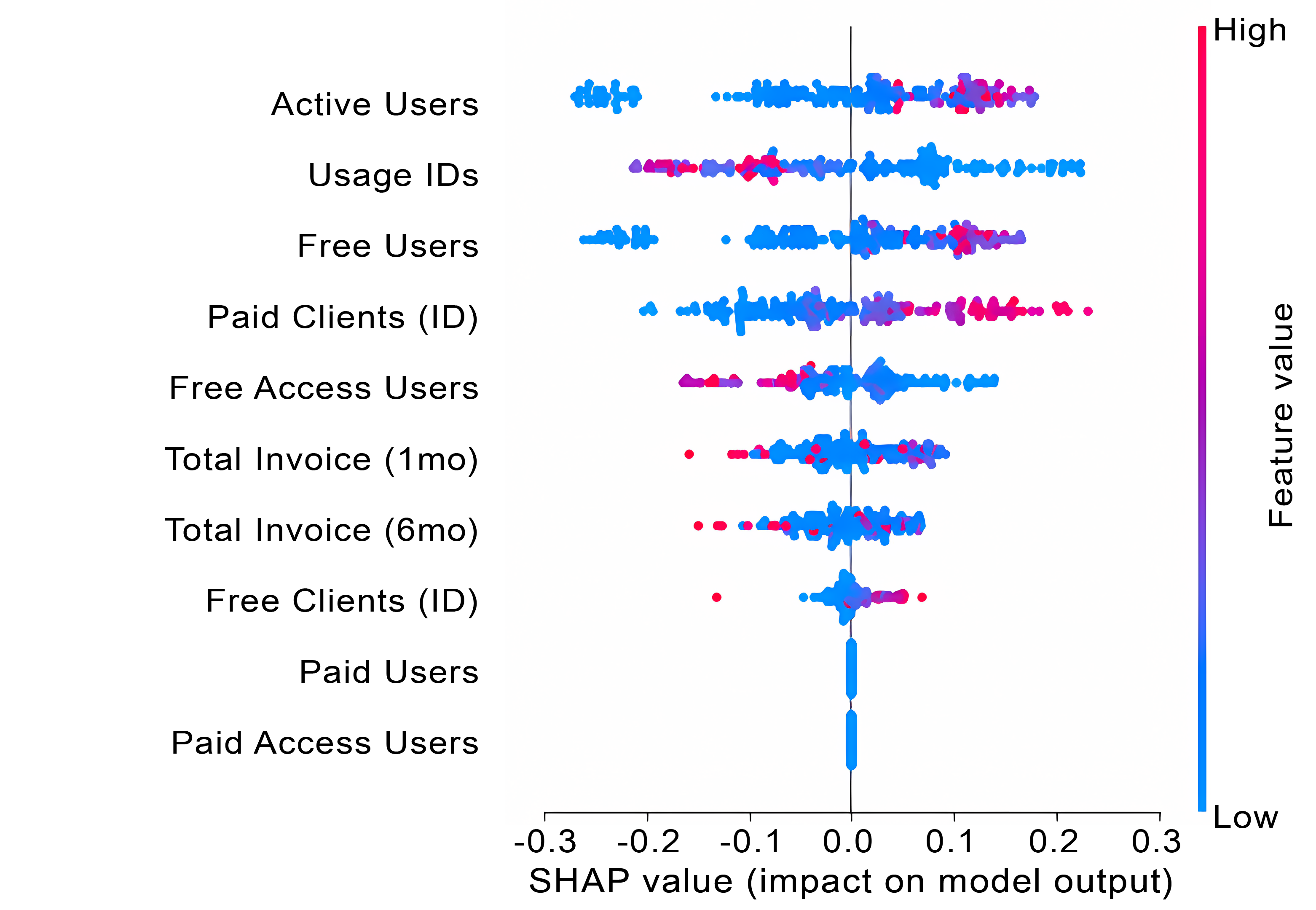}
    \caption{SHAP summary plot for real data (Enterprise Invoice).}
    \label{fig:inf_shap_real_summary}
  \end{subfigure}
  \hfill
  \begin{subfigure}{0.45\linewidth}
    \centering
    \includegraphics[width=\linewidth]{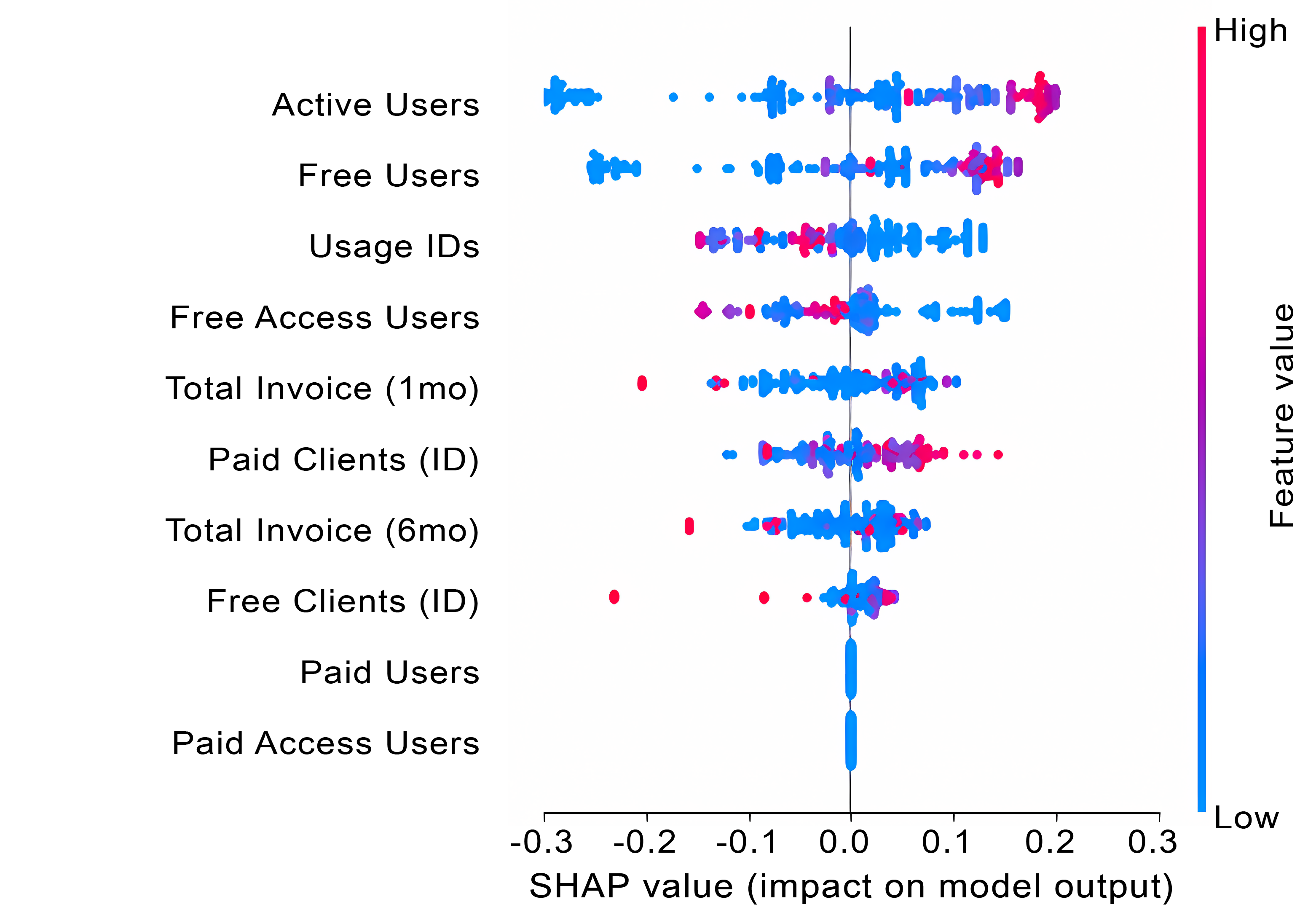}
    \caption{SHAP summary plot for synthetic data (Enterprise Invoice).}
    \label{fig:inf_shap_syn_summary}
  \end{subfigure}

  \vspace{0.5cm} 

  \begin{subfigure}{0.45\linewidth}
    \centering
    \includegraphics[width=\linewidth]{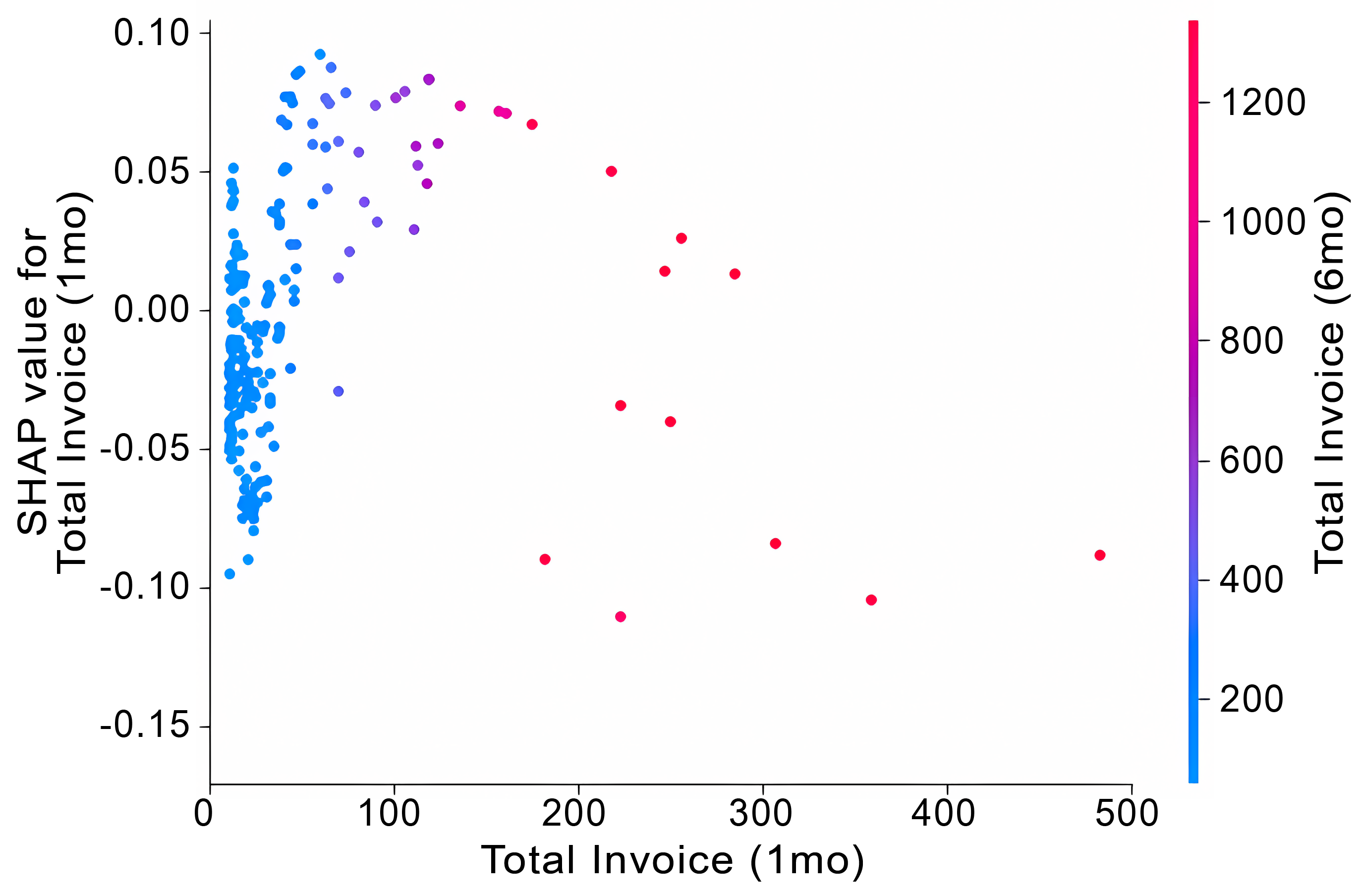}
    \caption{SHAP dependence plot for real data: effect of recent monthly invoices (colored by 6-month cumulative count).}
    \label{fig:inf_shap_real_dependence}
  \end{subfigure}
  \hfill
  \begin{subfigure}{0.45\linewidth}
    \centering
    \includegraphics[width=\linewidth]{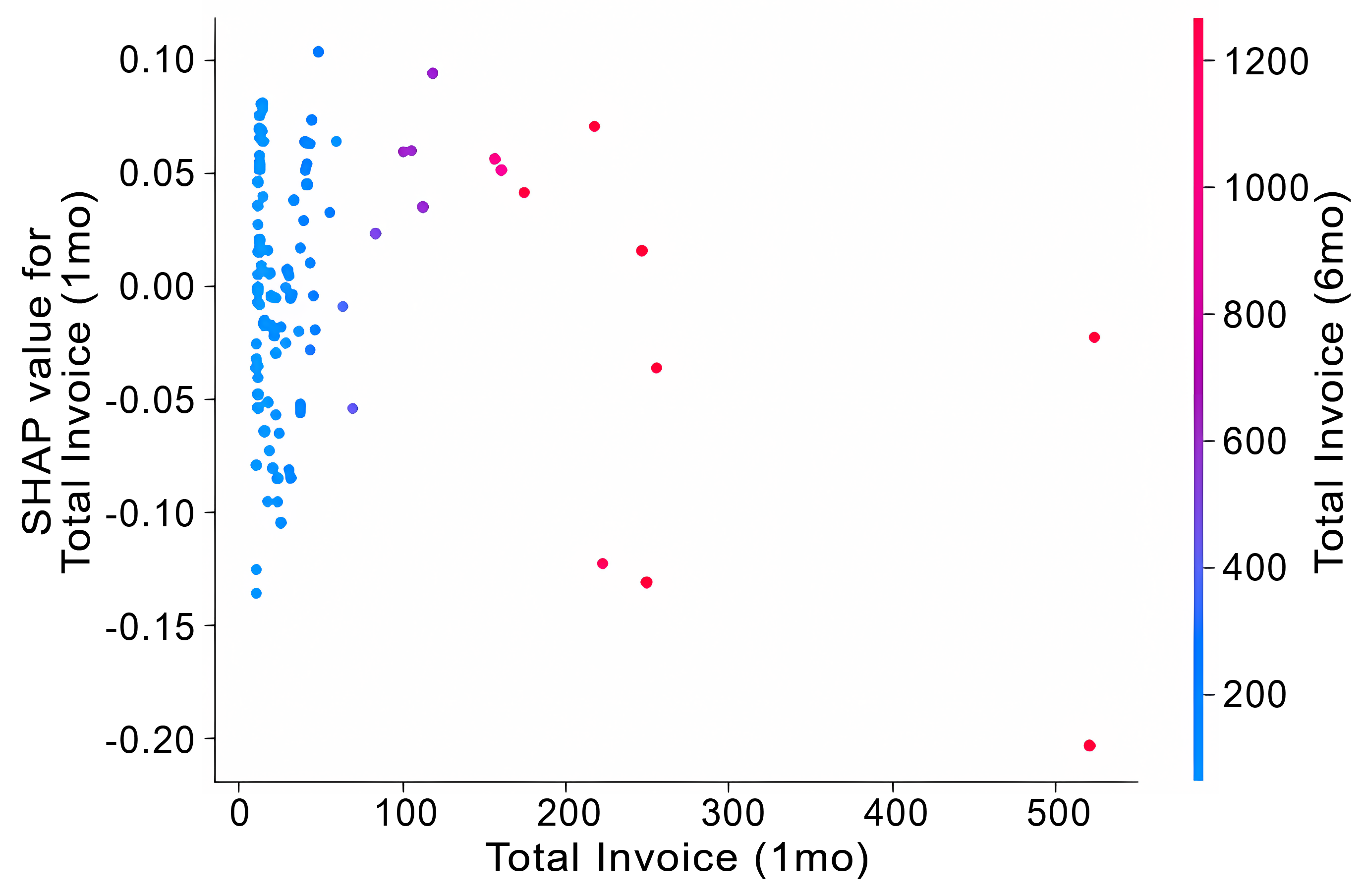}
    \caption{SHAP dependence plot for synthetic data: effect of recent monthly invoices (colored by 6-month cumulative count).}
    \label{fig:inf_shap_syn_dependence}
  \end{subfigure}

  \caption{SHAP-based feature attribution consistency analysis on the Enterprise Invoice dataset: 
  (a) summary plot (real data), 
  (b) summary plot (synthetic data), 
  (c) dependence plot for real data, and 
  (d) dependence plot for synthetic data.}
  \label{fig:inf_shap_overview}
\end{figure*}

\subsection{SHAP-Based Feature Attribution Consistency Analysis}

\begin{figure*}[htbp]
  \centering
  \begin{subfigure}{0.35\linewidth}
    \centering
    \includegraphics[width=\linewidth]{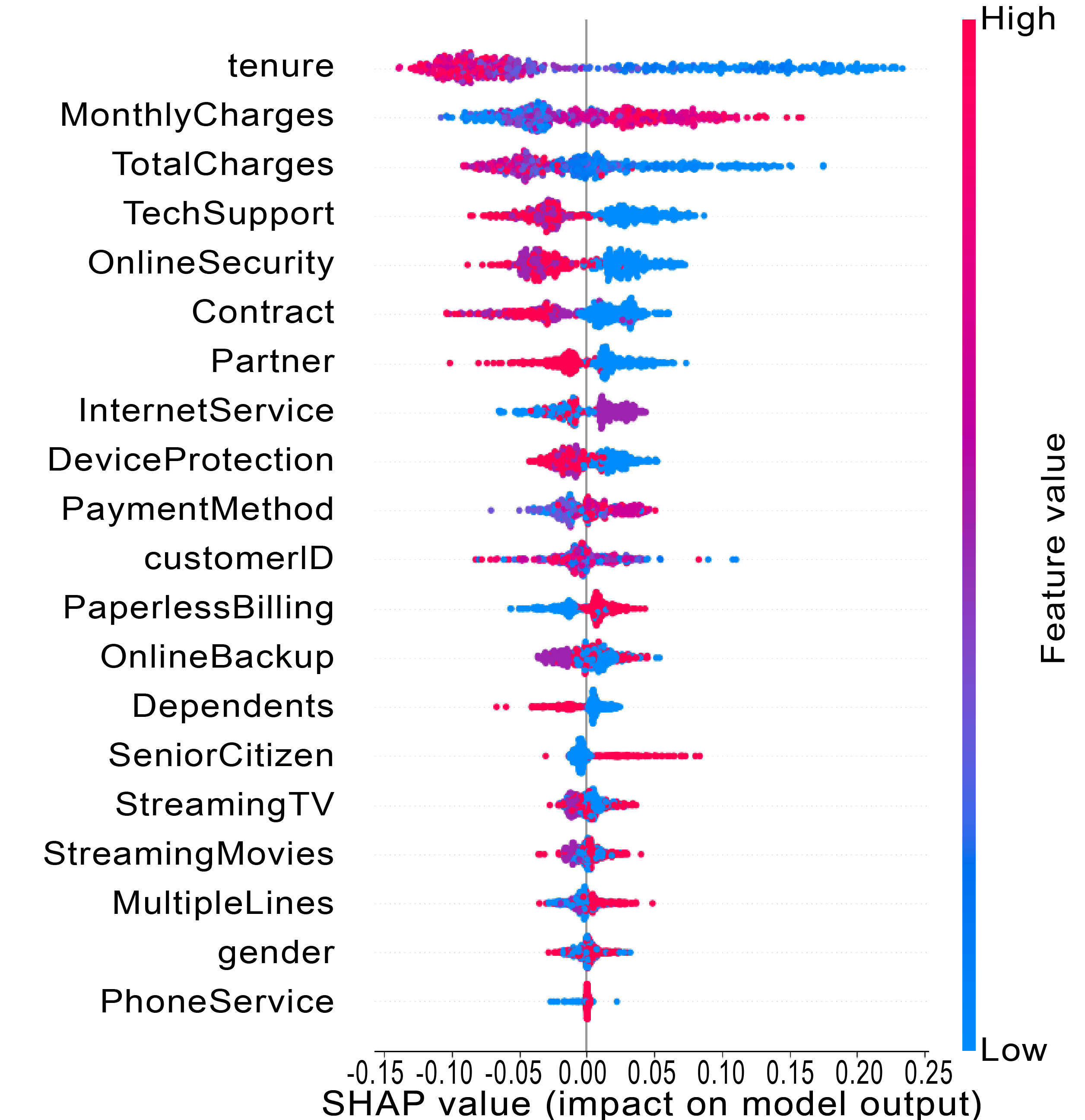}
    \caption{SHAP summary plot for real data (Telco Churn).}
    \label{fig:telco_shap_real_summary}
  \end{subfigure}
  \hfill
  \begin{subfigure}{0.35\linewidth}
    \centering
    \includegraphics[width=\linewidth]{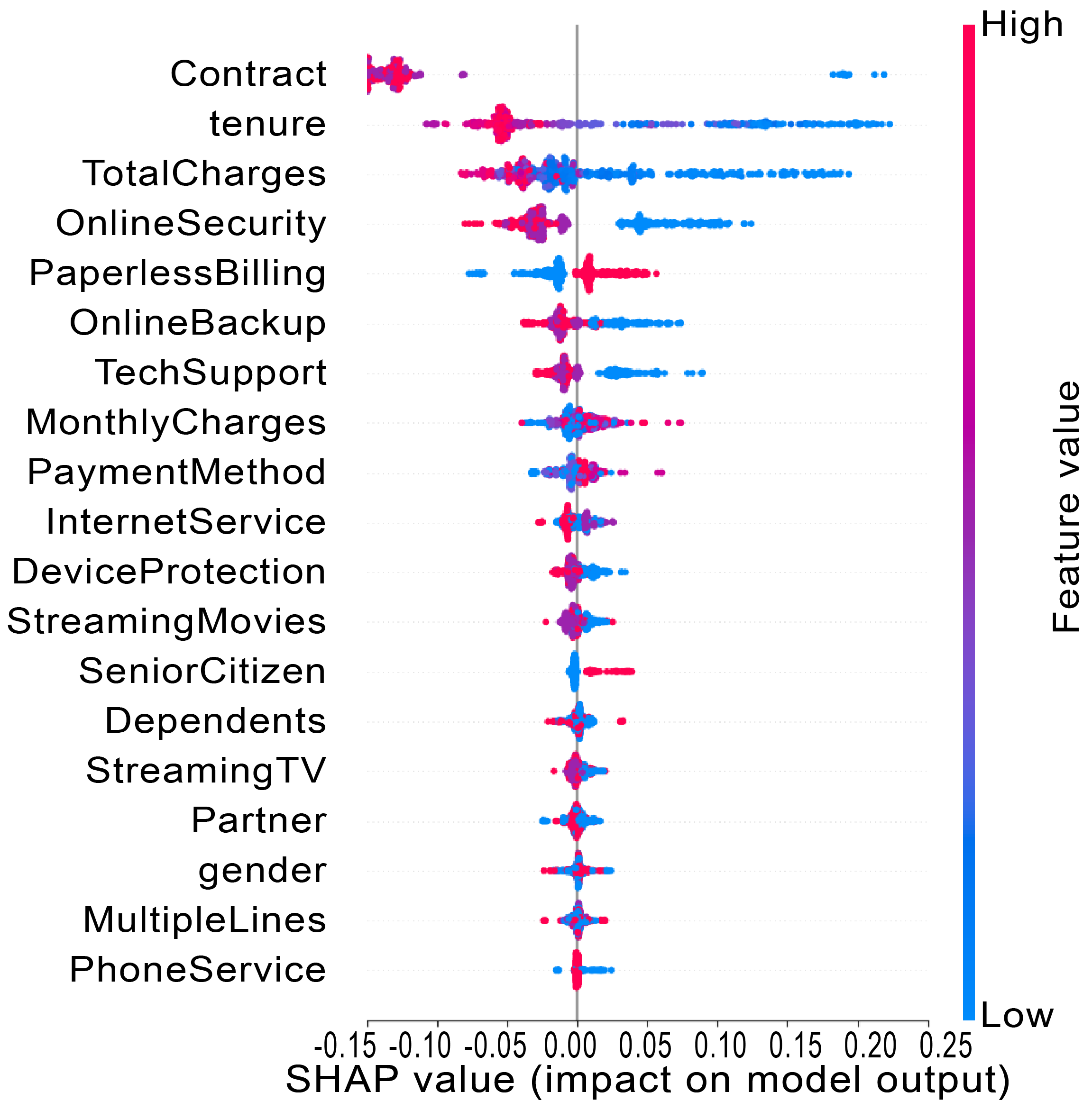}
    \caption{SHAP summary plot for synthetic data (Telco Churn).}
    \label{fig:telco_shap_syn_summary}
  \end{subfigure}

  \vspace{0.5cm} 

  \begin{subfigure}{0.45\linewidth}
    \centering
    \includegraphics[width=\linewidth]{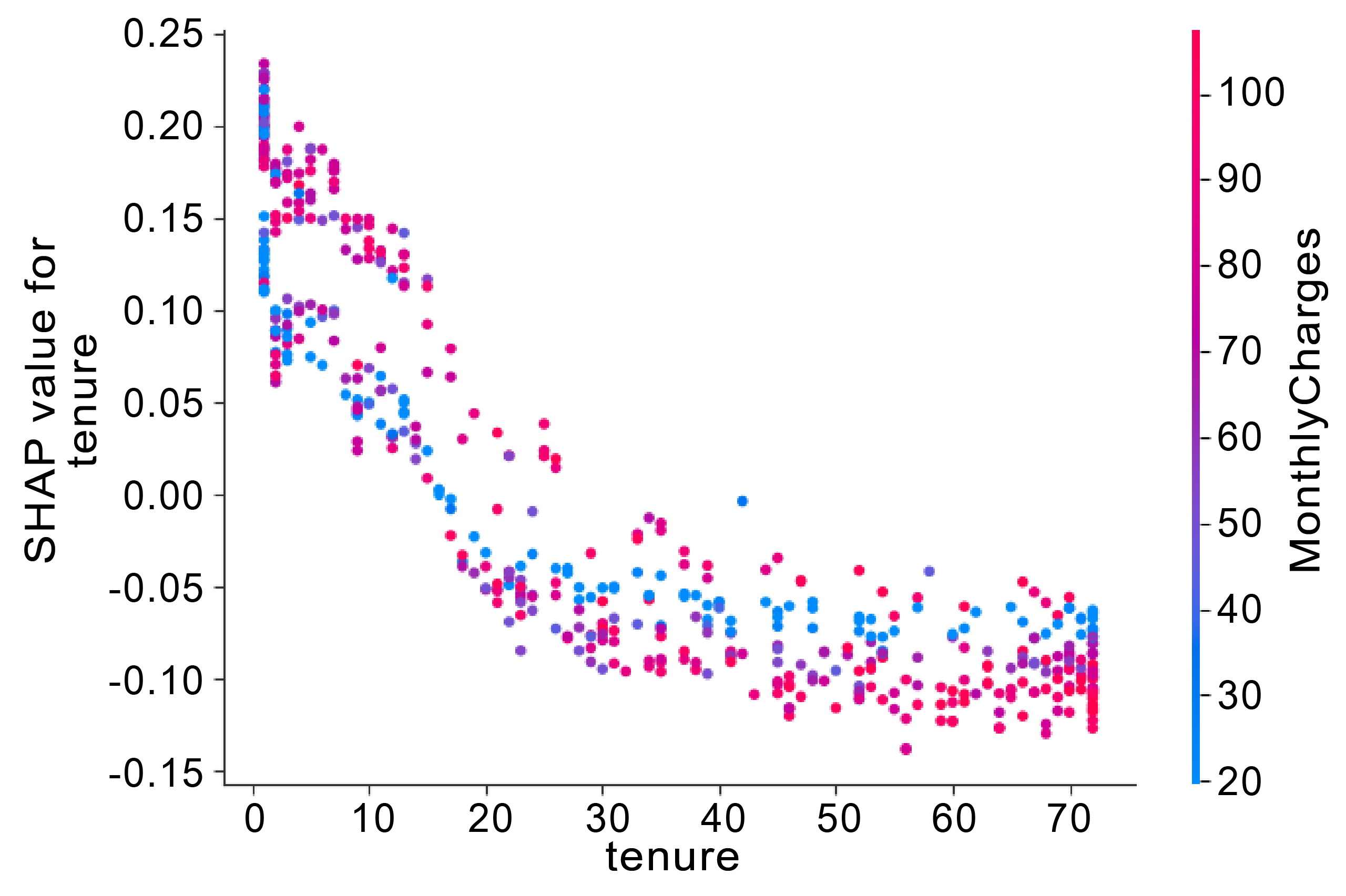}
    \caption{SHAP dependence plot for real data: effect of \texttt{tenure} (colored by \texttt{MonthlyCharges}).}
    \label{fig:telco_shap_real_dependence1}
  \end{subfigure}
  \hfill
  \begin{subfigure}{0.45\linewidth}
    \centering
    \includegraphics[width=\linewidth]{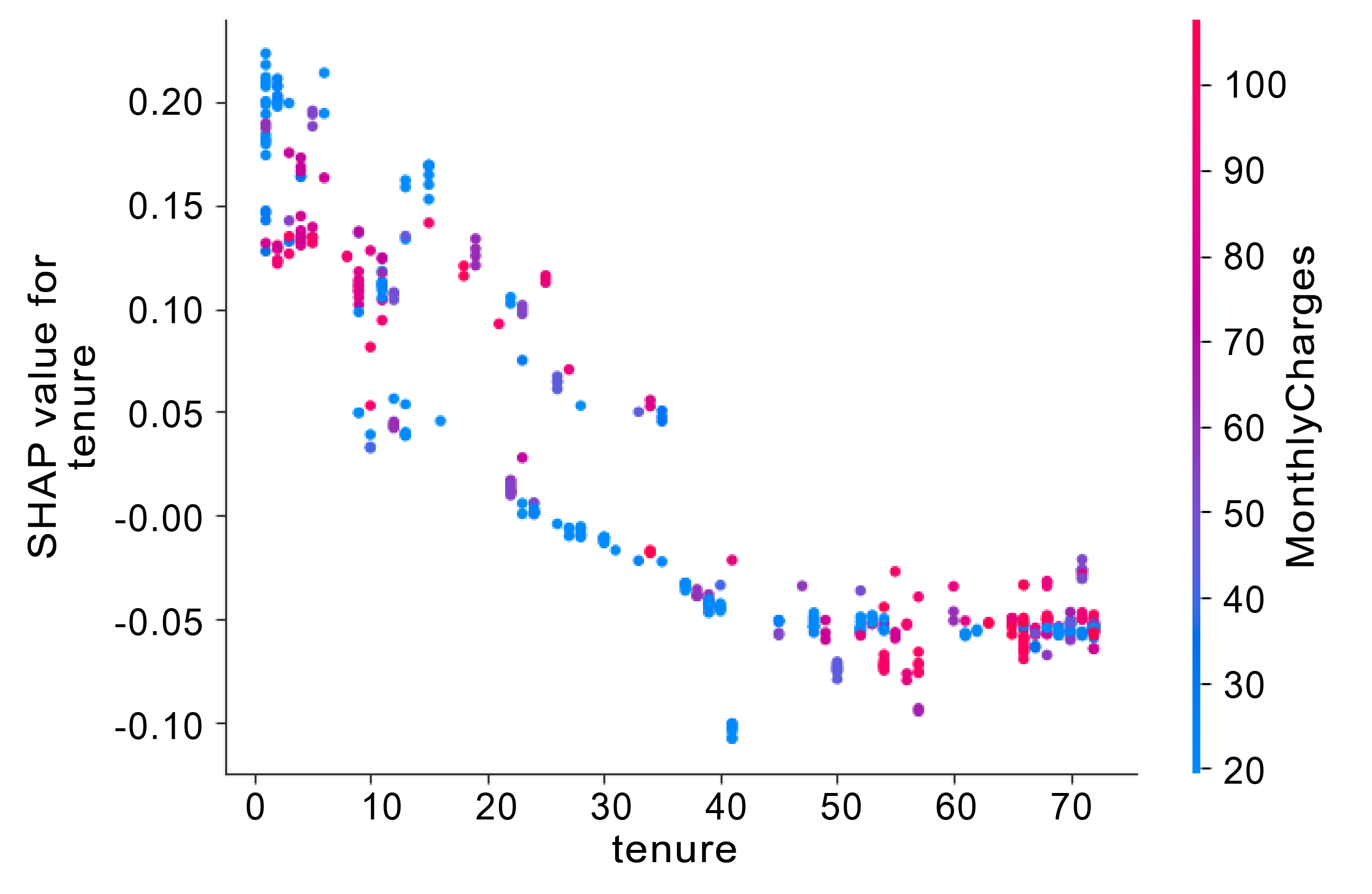}
    \caption{SHAP dependence plot for synthetic data: effect of \texttt{tenure} (colored by \texttt{MonthlyCharges}).}
    \label{fig:telco_shap_syn_dependence1}
  \end{subfigure}

  \caption{SHAP-based feature attribution consistency analysis on the Telco Churn dataset: 
  (a) summary plot (real data), 
  (b) summary plot (synthetic data), 
  (c) dependence plot for real data, and 
  (d) dependence plot for synthetic data.}
  \label{fig:telco_shap_overview}
\end{figure*}

To evaluate whether synthetic data preserves not only statistical properties but also the semantic decision logic embedded in downstream classifiers, we analyzed the SHAP values for models trained on both the real and synthetic datasets. This analysis highlights global attribution consistency, local interaction fidelity, and potential deviations arising from the generation process. Figures~\ref{fig:uci_shap_real_summary}--\ref{fig:telco_shap_syn_dependence1} show the results for the three datasets.

\textbf{UCI Heart Disease.}  
In the real dataset (Figures~\ref{fig:uci_shap_real_summary} and \ref{fig:uci_shap_real_dependence}), clinical features such as chest pain type (\texttt{cp}), number of major vessels (\texttt{ca}), and thalassemia (\texttt{thal}) dominate the model’s predictive behavior. The dependence plots reveal structured nonlinear interactions, e.g., between \texttt{age} and \texttt{chol}. The synthetic data (Figures~\ref{fig:uci_shap_syn_summary}, \ref{fig:uci_shap_syn_dependence}) broadly preserves this hierarchy, but with a flatter attribution profile, higher variability, and simplified interaction patterns. This indicates that while major decision pathways are retained, fine-grained nonlinearities are underrepresented.

\textbf{Enterprise Invoice Usage.}  
For the Infomart dataset (Figures~\ref{fig:inf_shap_real_summary} and \ref{fig:inf_shap_real_dependence}), usage-related features such as \texttt{NumUsers}, \texttt{NumIDs}, and \texttt{NumFreeUsers} emerged as key predictors. Their directional impact aligns with business intuition: higher usage increases the probability of invoice issuance. The synthetic data (Figures~\ref{fig:inf_shap_syn_summary} and \ref{fig:inf_shap_syn_dependence}) maintained the same top features and monotonic relationships, confirming the semantic preservation of business rules. However, attribution hierarchies appeared flatter and interaction effects (e.g., between one-month and six-month invoice counts) were smoother, suggesting that extreme-value dependencies are diluted.

\textbf{Telco Churn.}  
In the Telco Churn dataset (Figures~\ref{fig:telco_shap_real_summary} and \ref{fig:telco_shap_real_dependence1}), features such as \texttt{tenure}, \texttt{MonthlyCharges}, and \texttt{TotalCharges} dominated the model behavior, which is consistent with churn theory. The dependence plots revealed nonlinear relationships, including risk reduction with longer tenures. The synthetic data (Figures~\ref{fig:telco_shap_syn_summary} and \ref{fig:telco_shap_syn_dependence1}) largely preserves these trends but showed smoother dependence curves and flatter attribution distributions. Notably, \texttt{Contract} received a disproportionately high attribution in the synthetic model, reflecting the bias from prompt templates, whereas the billing-related features were comparatively underweighted.

\textbf{Cross-Dataset Trends.}  
Across all datasets, the synthetic data reproduced the global attribution hierarchy and major semantic relationships but exhibited two consistent limitations: (1) flattened attribution distributions that reduced the dominance of the top features, and (2) smoother or simplified interaction effects that underrepresented tail behaviors and nonlinear dependencies. These discrepancies emphasize the value of SHAP-based analysis for diagnosing semantic fidelity gaps that remain hidden under purely statistical evaluations.

\subsection{Statistical Gap Overview}

To provide a unified overview of the statistical discrepancies between real and synthetic data, 
we report the dataset-wide gap metrics across UCI Heart Disease, Enterprise Invoice, and Telco Churn. 
The results in Table~\ref{tab:statistical_gaps} summarizes the mean gap, standard deviation gap, covariance gap, 
and Spearman rank correlation, capturing the alignment in central tendency, variability, multivariate dependence, 
and feature rank-order consistency. 

The UCI Heart Disease dataset showed moderate mean and covariance gaps, whereas the Enterprise Invoice dataset exhibited the largest 
covariance discrepancy owing to heterogeneous feature scales and dense correlation structures. 
The Telco Churn dataset demonstrates the strongest statistical alignment, with the lowest covariance gap 
and highest Spearman correlation coefficient (0.8702), although the mean gap remained relatively large. 
These observations highlight that, across datasets, the synthetic data were not perfectly matched to the real data 
at the distributional level, particularly in terms of higher-order covariance.

Nevertheless, the key finding is that even with these statistical deviations, 
synthetic data refined using the SHAP Distance achieved the highest downstream utility.
This indicates that semantic fidelity, as measured by attributional consistency, is more important than
perfect statistical resemblance when the ultimate goal is to support predictive modeling and 
real-world decision-making.

\begin{table}[h]
\centering
\caption{Statistical gap metrics across datasets}
\label{tab:statistical_gaps}
\resizebox{\linewidth}{!}{%
\begin{tabular}{lcccc}
\toprule
\textbf{Dataset} & \textbf{Mean Gap} & \textbf{Std. Gap} & \textbf{Cov. Gap} & \textbf{Spearman Corr.} \\
\midrule
UCI Heart Disease     & 2.8780   & 12.8360   & 2,509.05     & 0.7692 \\
Enterprise Invoice    & 19.6455  & 143.5829  & 3,466,926.39 & 0.6951 \\
Telco Churn           & 24.9491  & 30.7742   & 558,392.03   & 0.8702 \\
\bottomrule
\end{tabular}%
}
\end{table}

\section{Conclusion}
In this work, we have proposed the SHAP Distance as a semantic fidelity metric for evaluating synthetic tabular data. Although attribution-based metrics such as the SHAP Distance provide a direct lens into reasoning consistency, they complement rather than replace other semantic evaluation approaches. For example, methods based on mutual information quantify how much predictive signal is preserved at the feature level, whereas counterfactual-consistency tests evaluate whether synthetic data induces stable decision outcomes under minimal perturbations. Compared to these approaches, SHAP Distance focuses on model-internal attribution alignment, providing a semantic perspective on whether synthetic data induces similar feature-usage patterns. Integrating these complementary perspectives could form a more holistic semantic evaluation framework for synthetic tabular data. Through experiments on three diverse datasets, we demonstrated that, while the synthetic data often exhibited noticeable deviations from the real data at the statistical level, such as marginal distributions and covariance structures, our SHAP-based evaluation highlighted stronger preservation of decision logic. More importantly, the synthetic data generated and refined with the SHAP Distance consistently achieved the highest downstream utility, indicating that semantic alignment, rather than purely statistical similarity, is the key to practical usefulness. This suggests that future research on synthetic tabular data should emphasize attribution-based consistency to ensure that the generated samples are not only statistically plausible but also semantically faithful and effective for downstream applications.

\section{Future Work}
Although the SHAP Distance provides a principled means of measuring semantic fidelity, several issues remain for future research. First, extending the metric to incorporate causal feature attributions would allow the evaluation of whether synthetic data preserve not only associations but also the underlying causal structures. Second, combining the SHAP Distance with attention mechanisms from transformer-based models can offer richer insights into feature interactions, particularly in high-dimensional datasets. Third, integrating semantic fidelity metrics into the generative loop opens up possibilities for automated refinement frameworks, where attribution misalignments dynamically guide the generation process. Finally, applying the SHAP Distance to multimodal or temporal datasets—such as electronic health records or financial transaction streams—could broaden its utility and reveal domain-specific challenges. These directions will help to establish semantic alignment as a central objective in synthetic data generation, ensuring that future models will produce data that are both realistic and truly useful for downstream tasks.

\section*{Acknowledgment}

This study was supported by the joint research project with Infomart Corporation and JST PRESTO (JPMJPR2369).

\end{document}